\documentclass[utf8, prx, superscriptaddress, twocolumn]{revtex4-2} \usepackage[ngerman,english]{babel}
\usepackage[utf8]{inputenc}
\usepackage{url,lineno,microtype}
\usepackage[detect-all,binary-units,separate-uncertainty=true]{siunitx}
\usepackage{tikz}
\usepackage{todonotes}
\usepackage{glossaries}
\usepackage{booktabs}
\usepackage{changes}
\usepackage{textcomp}

\PassOptionsToPackage{hyphens}{url}\usepackage[hidelinks]{hyperref}
\usepackage{cleveref}

\DeclareSIUnit{\nothing}{\relax}

\newacronym{sram}{SRAM}{static random access memory}
\newacronym{simd}{SIMD}{single-instruction multiple data}
\newacronym{dac}{DAC}{digital-to-analog converter}
\newacronym{adc}{ADC}{analog-to-digital converter}
\newacronym{cadc}{CADC}{column analog-to-digital converter}
\newacronym{asic}{ASIC}{application-specific integrated circuit}
\newacronym{ppu}{PPU}{plasticity processing unit}
\newacronym{adex}{AdEx}{adaptive exponential integrate-and-fire}
\newacronym{lif}{LIF}{leaky integrate-and-fire}
\newacronym{snn}{SNN}{spiking neural network}
\newacronym{stdp}{STDP}{spike timing-dependent plasticity}
\newacronym{stp}{STP}{short-term plasticity}
\newacronym{epsp}{EPSP}{excitatory post synaptic potential}
\newacronym{bptt}{BPTT}{backpropagation through time}
\newacronym{fpga}{FPGA}{Field Programmable Gate Array}

\DeclareSIUnit\op{Op}
\begin{document}
\title[The BSS-2 accelerated neuromorphic system with hybrid plasticity]{The BrainScaleS-2 accelerated neuromorphic system with hybrid plasticity} 

\author{Christian Pehle}
\thanks{These authors contributed equally}
\affiliation{Kirchhoff-Institute for Physics, Heidelberg University, Germany}
\author{Sebastian Billaudelle}
\thanks{These authors contributed equally}
\affiliation{Kirchhoff-Institute for Physics, Heidelberg University, Germany}
\author{Benjamin Cramer}
\thanks{These authors contributed equally}
\affiliation{Kirchhoff-Institute for Physics, Heidelberg University, Germany}
\author{Jakob Kaiser}
\thanks{These authors contributed equally}
\affiliation{Kirchhoff-Institute for Physics, Heidelberg University, Germany}
\author{Korbinian Schreiber}
\thanks{These authors contributed equally}
\affiliation{Kirchhoff-Institute for Physics, Heidelberg University, Germany}
\author{Yannik Stradmann}
\thanks{These authors contributed equally}
\affiliation{Kirchhoff-Institute for Physics, Heidelberg University, Germany}
\author{Johannes Weis}
\thanks{These authors contributed equally}
\affiliation{Kirchhoff-Institute for Physics, Heidelberg University, Germany}
\author{Aron Leibfried}
\affiliation{Kirchhoff-Institute for Physics, Heidelberg University, Germany}
\author{Eric Müller}
\affiliation{Kirchhoff-Institute for Physics, Heidelberg University, Germany}
\author{Johannes Schemmel}
\email{schemmel@kip.uni-heidelberg.de}
\affiliation{Kirchhoff-Institute for Physics, Heidelberg University, Germany}
\begin{abstract} 
Since the beginning of information processing by electronic components, the nervous system has served as a metaphor for the organization of computational primitives. Brain-inspired computing today encompasses a class of approaches ranging from using novel nano-devices for computation to research into large-scale neuromorphic architectures, such as TrueNorth, SpiNNaker, BrainScaleS, Tianjic, and Loihi. While implementation details differ, spiking neural networks -- sometimes referred to as the third generation of neural networks -- are the common abstraction used to model computation with such systems. Here we describe the second generation of the BrainScaleS neuromorphic architecture, emphasizing applications enabled by this architecture. It combines a custom analog accelerator core supporting the accelerated physical emulation of bio-inspired spiking neural network primitives with a tightly coupled digital processor and a digital event-routing network.
 \end{abstract}
    
\maketitle

\section{Introduction}
One important scientific goal of computational neuroscience is the advancement of brain-inspired computing. Continuous-time emulators for modeling brain function play an essential role in this endeavor. They provide resource-efficient platforms for the bottom-up modeling of brain function -- including computationally expensive aspects like plasticity and learning or structured neurons.
BrainScaleS is a neuromorphic computing platform that realizes this approach to the furthest extent possible with current technologies by constructing a physical replica of the most commonly used reductionist view of the biological brain: a network of neurons connected via plastic synapses.

In this aspect, it differs from most other modeling approaches within the computational neuroscience community. In particular, while the network model operates, no differential equation gets solved. Biological processes are not represented by discrete-time changes of a multitude of bits representing some binary approximation of molecular biology. Instead, the temporal evolution of physical quantities, such as current and voltage, directly correspond to the neural dynamics in BrainScaleS. In that regard, our approach is similar to system architectures using novel nano-devices to perform computation. However, we focus on creating a controllable,  configurable substrate based on well-understood CMOS technology, which can serve as a platform for research into system-level aspects of such an approach.

In designing the BrainScaleS-2 architecture, we had several high-level design goals and use cases in mind. Some of them are informed by the limitations we discovered with previous system designs. The overarching design goal was to enable large-scale accelerated emulation of spiking neural networks. This requires a \emph{scalable} system architecture. The system we will describe in this article is a \emph{unit of scale} for such a large-scale system. Analog neuromorphic hardware presents unique challenges to scalability compared to digital neuromorphic hardware. Since the constituting components are subject to both fixed-pattern noise, as well as temperature and time-dependent drift of parameters, architectural solutions are needed to address these. Device variations (fixed-pattern noise) are addressable by calibration, but this requires acceptable parameter ranges and sufficient compute resources, which enable constant scaling with system size. A novel design for analog parameter storage \citep{hock13analogmemory} ensures wide parameter ranges and stability of parameters over time. Rapid calibration is enabled by including embedded processor cores and high-resolution and multi-channel analog-to-digital converters with access to all relevant analog states. Taken together, they enable \emph{on-chip calibration} in our unit of scale and ensure that calibration time remains constant with system size.

From a user perspective of the system, we aim to support several operation modes and use cases for the system architecture. They can be distinguished along several dimensions. Perhaps the most common mode of operation is to perform experiments in \emph{batch mode}. In this mode, experiment instances are queued and sequentially executed on the system without any data dependency among the different instances. Batch-mode execution is used for parameter sweeps, evaluation of biology-inspired learning rules, or inference once task-specific parameters for a classification task have been found. There are several ways to introduce data dependencies between experiment instances. We refer to them collectively as \emph{in-the-loop} operation. Our system enables us to close these loops on several time scales and at different points of the system hierarchy, as we will further describe in \cref{ssec:hybrid_plasticity_and_versatile_digital_control}. This ability features also prominently in some of the experiments: among them the learning-to-learn approach \citep{bohnstingl2019neuromorphic}, briefly discussed in \cref{ssec:biology_inspired_learning_approaches}, the surrogate gradient in-the-loop optimization described in \cref{ssec:surrogate_gradient_itl}, and analog artificial neural network training in \cref{ssec:gradient_based_ann}.

One of the key distinguishing features of the architecture is our approach to synaptic plasticity. In contrast to designs like Loihi \citep{davies2018loihi}, which only supports micro-coded operations per synapse or other designs with fixed plasticity, we support plasticity programs with complex control and data dependencies. The combination of massive-parallel data acquisition of (analog) system observables (synaptic correlation and membrane voltage traces) and the efficient, digital evaluation in programmable plasticity rules is a unique strength of our system. This approach will be described in more detail in \cref{ssec:hybrid_plasticity_and_versatile_digital_control} and is the basis of all experiments reported in \cref{ssec:biology_inspired_learning_approaches}.

Another key distinguishing feature is that our synaptic crossbar can process weighted spikes. This capability enables the use of the same components to implement analog vector-matrix multiplication. We will give an overview of the analog emulation of artificial neural networks on our system in \cref{ssec:accelerated_emulation}.

The rest of this paper is structured as follows: We begin with a more detailed description of the system architecture in \cref{ssec:system_architecture}. Afterwards, we describe the analog core of the system in \cref{ssec:accelerated_emulation}, in particular the design of the neuron circuitry with an emphasis on the design decisions that lead to controllability and a wide range of biological parameters. Beyond the analog core, the system also incorporates two loosely coupled embedded processors enabling the realization of \emph{hybrid plasticity} schemes, emulation of virtual environments for reinforcement experiments, as well as the orchestration of calibration and data transfer. We describe this part of the system in \cref{ssec:hybrid_plasticity_and_versatile_digital_control}.

Taken together, these design decisions and features of the system architecture enable the use of the system on a wide range of tasks and operation modes. The current system design can serve as a versatile platform for experimentation with both biology-inspired and machine-learning-inspired learning approaches. We will present experiments supporting these assertions in \cref{ssec:biology_inspired_learning_approaches,ssec:gradient_based_learning_approaches}. From a system design perspective, we see these results as evidence that the design is suitable as a unit of scale for a large-scale accelerated neuromorphic learning architecture.

Finally, we will discuss related work and give an outlook on future developments in \cref{sec:discussion}. \section{The BrainScaleS-2 system}
\subsection{System Architecture}
\label{ssec:system_architecture}
\begin{figure*}[!ht]
    \begin{center}
    \includegraphics[width=\textwidth]{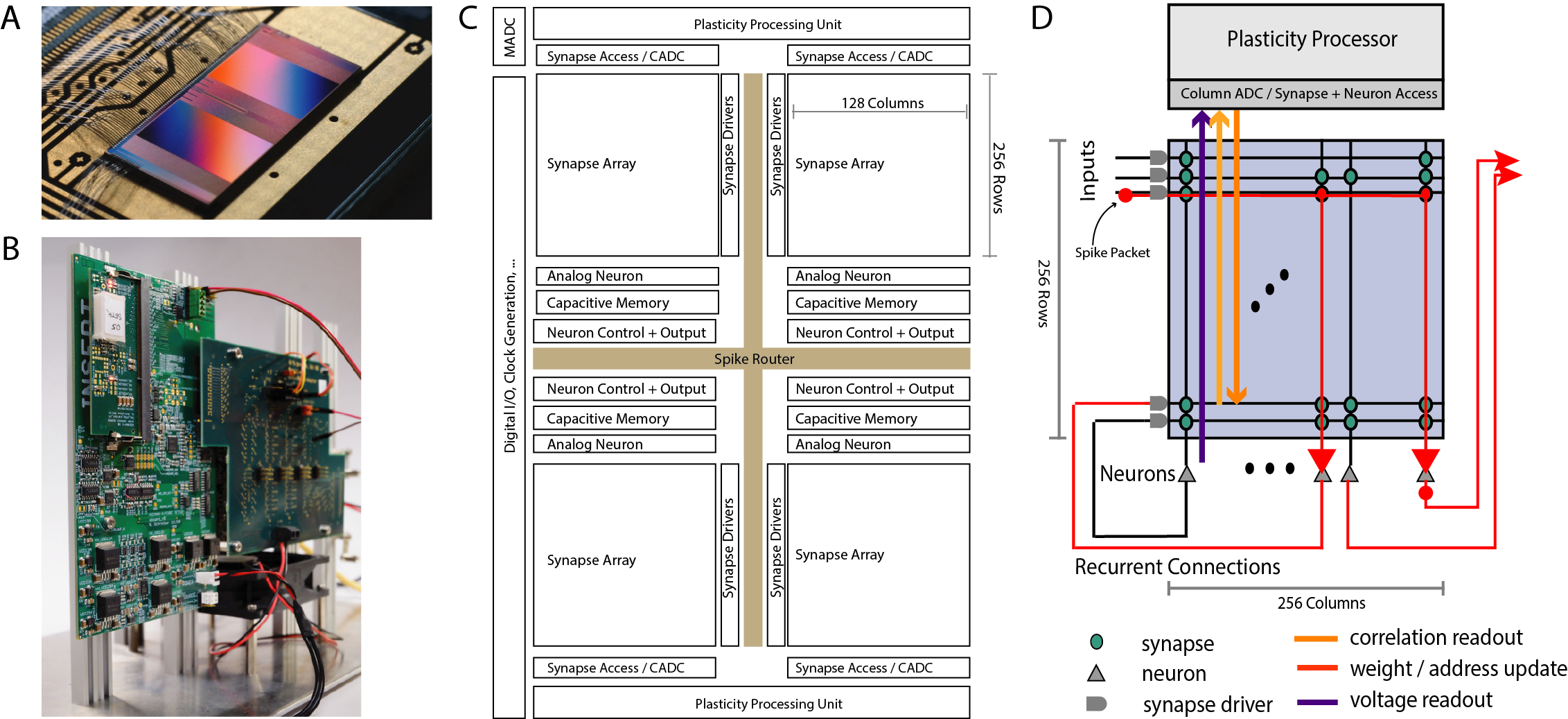}
    \end{center}
	\caption{Overview of the BrainScaleS-2 System architecture. (A) Bonded chip on its carrier board, one can see the two synaptic crossbar arrays. (B) Test setup, with the chip (covered by white plastic) mounted on a carrier board. The \gls{fpga} and I/O boards have been designed by our collaboration partners at TU Dresden. (C) Schematic floorplan of the chip: Two processor cores with access to the synaptic crossbar array are on the top and bottom. The \num{512} neuron circuits and analog parameter storage are arranged in the middle. The event router routes events generated by the neurons and external events to the synapse drivers and to/from the digital I/O located on the left edge of the chip. (D) Conceptual view of the system architecture in spike processing mode: Event packets (red dot) get injected by the synapse driver into the synaptic crossbar, where they cause synaptic input integration to occur in synapses with matching addresses (indicated by red lines). Membrane voltage accumulation eventually results in spike generation in the associated neuron circuits. The resulting spikes are routable to both synapse drivers or external output. The plasticity processing unit has low latency and massively parallel access to synaptic weights, addresses, correlation measurements, and neuron membrane voltage dynamics during operation. Plasticity rules and other learning algorithms can use these observables to modify all parameters determining network emulation in an online fashion.}
    \label{fig:system_architecture}
\end{figure*}

This section will give an overall description of the BrainScaleS-2 architecture in terms of its constituting components. We have taped out several scaled-down prototype versions and now successfully commissioned a full-scale single-core system. This single-core system can serve as the unit of scale for larger-scale designs involving multiple neuromorphic cores. In \cref{fig:system_architecture} we show an overview of the system architecture.

A single neuromorphic BrainScaleS-2 core consists of a full-custom analog core combining a synaptic crossbar, neuron circuits, analog parameter storage, two digital control- and plasticity-processors, and the event routing network responsible for spike communication. The physical design is divided into four quadrants, each featuring a synaptic crossbar with \num{256} rows and \num{128} columns. The neuron circuitry is also partitioned in this way, with each of the \num{512} neuron circuits associated with one column. The two digital processors \citep{friedmann2016hybridlearning} located at the top and bottom of the design are responsible for the analog core's upper and lower half, respectively. Using \gls{simd} vector extensions, they can read and write the digital state of their half of the synaptic crossbars row-wise in parallel and readout analog traces via a \num{512}-channel \gls{cadc} (there are \num{256} channels per quadrant, pairwise associated to correlation and anti-correlation measurement). In \cref{ssec:hybrid_plasticity_and_versatile_digital_control} we will give a more detailed description of the digital processors' role in the realization of plasticity. The event routing network takes up the cross-shaped space dividing the four quadrants and is extendable in all directions. Operating on address event-encoded packets, it connects the digital output of neuron circuits, external input, and on-chip Poisson sources and routes them to synaptic rows of one of the quadrants or off-chip targets. The digital interface of events and data to an external system implements a custom hardware link protocol \citep{karasenko2020arq}, which supports both reliable data transfer and efficient event communication.

The single-core BrainScaleS-2 system has been integrated into two hardware setups so far: One system tailored for commissioning and analog measurements (shown in \cref{fig:system_architecture} B). In addition, we designed and implemented a fully integrated ``mobile'' system meant for edge deployment, which incorporates a Zynq \gls{fpga} with embedded ARM cores that can run Linux and the BrainScaleS-2 neuromorphic core in a small form-factor \citep{stradmann2021demonstrating}. It, therefore, can be deployed entirely independently from any external host. In both cases the \gls{fpga} is used for real-time control, buffering of external stimulus and output data. It also manages external memory access for the plasticity processors and provides connection from and to the host system. 

Beyond these two hardware setups, we also see the single-core system as a \emph{unit of scale} for large-scale neuromorphic architectures. The most immediate step that does not require a modified ASIC architecture is integrating multiple single-core chips into a larger system by implementing an external event-routing architecture. Work in this direction is underway. We can also increase the reticle size and extend the on-chip digital routing architecture appropriately. With this approach, about \num{8} cores could be integrated on a single reticle. We will discuss some future directions in \cref{sec:discussion}.

Further improvements might consider modifications to the core architecture itself. Future revisions will include on-chip memory controllers, thereby eliminating the need to rely on an external \gls{fpga}. The overall modularity of the architecture would also allow swapping out the synaptic crossbar, neuron circuit, or plasticity processor implementation. The neuron circuit implementation underwent several iterations until it converged on the design reported here. We will turn to this aspect of the system next and then describe the plasticity and control processor in more detail.
 \subsection{Accelerated Analog Emulation of Neural Dynamics}
\label{ssec:accelerated_emulation}
The emulation of neuron and synapse dynamics takes place in dedicated mixed-signal circuits, which -- in combination with other full-custom components -- constitute the analog neuromorphic core.
A full-sized \gls{asic} encompasses \num{512} neuron compartments with versatile and rich dynamics.
They evolve at \num{1000}-fold accelerated time scales compared to the biological time domain, paying tribute to the characteristic time constants of the semiconductor substrate.
In its core, the neuron circuits faithfully implement the \gls{adex} model~\citep{brette2005adaptive}
\begin{align}
	C_\text{m} \dot{V} &= -g_\text{l} (V - E_\text{l}) + g_\text{l} \Delta_\text{T} \exp\left(\frac{V - V_\text{T}}{\Delta_\text{T}}\right) - w + I \,, \label{eq:adex}\\
	\tau_\text{w} \dot{w} &= a (V - E_\text{L}) - w \,,
\end{align}
where the first differential equation describes the evolution of the membrane potential $V$ on capacitance $C_\text{m}$.
The membrane accumulates currents $I = I_\text{syn} + I_\text{stim}$, which encompass direct external stimuli as well as currents originating from synaptic interaction.
Furthermore, $g_\text{l}$ represents the leak conductance pulling the membrane towards the leak potential $E_\text{l}$.
The exponential term implements the strong positive feedback emulating the coarse shape of the action potential of a biological neuron and is controlled by the exponential slope $\Delta_\text{T}$ and the soft threshold $V_\text{T}$.
An outgoing spike is released as soon as the membrane potential crosses the hard firing threshold $V_\text{th}$.
In that case, the membrane is clamped to the reset potential $V_\text{r}$ and held there for the refractory period $t_\text{r}$.
A second differential equation captures the dynamics of the adaptation current $w$ allowing the neuron to adapt to its previous activation and firing activity.
The adaptation state decays back to zero with a time constant $\tau_\text{w}$ and is driven by the deflection of the membrane potential, scaled with the subthreshold adaptation strength $a$.
In case of an action potential, $w$ is incremented by $b$ implementing spike-triggered adaptation.
A more detailed, transistor-level description of a previous version of the BrainScaleS-2 neuron circuit can be found in~\citet{aamir2018dls2neuron}.

Each neuron circuit can be configured individually via \SI{80}{\bit} of local configuration \gls{sram} as well as \num{24} analog parameters which are provided by an on-chip \gls{dac} with \SI{10}{\bit} resolution~\citep{hock13analogmemory}.
The analog parameters allow to control all potentials and conductances mentioned in \cref{eq:adex} for each neuron individually -- the model dynamics can therefore be tuned precisely and production-induced fixed-pattern deviations can be compensated.
This in particular allows to calibrate each circuit to a specific set of model parameters, which may either be homogeneous across the whole array or custom to individual neuron instances.
Other aspects of the neuron, such as the refractory time $t_\text{r}$ or the membrane capacitance $C_\text{m}$, can be directly configured via the locally stored digital configuration, which can also be used to en- or disable certain features of the neuron.
For example, the adaptation current as well as the exponential term in \cref{eq:adex} can be disconnected from the membrane, reducing the \gls{adex} model to the simpler \gls{lif} neuron model.
When disabling also the leak and threshold circuits, the neuron can be employed to linearly accumulate charges and therefore -- in conjunction with the synapse array -- implement analog matrix multiplication (see~\cref{ssec:gradient_based_ann}).

On the other hand, the neuronal dynamics can also be extended.
Using additional resistors and switches between the neuron circuits, larger cells with an increased synaptic fan-in as well as intricately structured neurons can be formed~\citep{aamir2018mixed,kaiser2021emulating}.

Each neuron circuit integrates synaptic stimuli from a column of \num{256} plastic synapses.
BrainScaleS-2, in particular, features time-continuous current- as well as conductance-based synapses with exponentially decaying kernels.
The total current
\begin{align}
	I_\text{syn} &= \sum_i w_i S_i(t) \ast \operatorname{e}^{-t/\tau_\text{syn}} \label{eq:i_syn}
\end{align}
hence results from the sum over all associated synapses $i$, their respective weights $w_i$, the presynaptic spike trains $S_i(t) = \sum_j \delta(t - t_j)$, and the synaptic time constants $\tau_\text{syn}$, which can be chosen independently for excitatory and inhibitory stimuli.
On BrainScaleS-2, the weights are stored locally per synapse in \SI{6}{\bit} \glspl{sram} and modulate the amplitude of an emitted current pulse.
Along them, each synapse also holds a \SI{6}{\bit} source address which is compared to the label of afferent events and lets the synapse then only responds to matching stimuli.
Hence, the network structure is determined not only through the digital event routing network but also by synapse-local properties.
Synapse addresses, in particular, allows to map sparse networks efficiently and change the connectome by inserting and removing synaptic connections dynamically~\citep{billaudelle2021structural}.

\Gls{stdp} and related correlation-based plasticity rules are supported through analog sensor circuits within each synapse~\citep{friedmannschemmel2016}.
They continuously measure the exponentially decaying pair-wise correlation between post- and presynaptic spikes and accumulate them as an observable for weight update calculations (cf.~\cref{ssec:hybrid_plasticity_and_versatile_digital_control}).
In addition, BrainScaleS-2 supports a presynaptic modulation of events, which is exploitable for the implementation of \gls{stp}~\citep{poo98stdp} and allows to inject graded spikes.
By combining the latter with neurons with disabled spiking dynamics -- hence acting as simple integrator circuits -- BrainScaleS-2 also supports the execution of non-time-continuous vector-matrix multiplications (see~\cref{ssec:gradient_based_ann}).
 \subsection{Hybrid Plasticity and Versatile Digital Control}
\label{ssec:hybrid_plasticity_and_versatile_digital_control}
Besides the accelerated and faithful emulation of neuron and synapse dynamics realized by the analog neuro-synaptic core described in the previous section, the system features two digital plasticity and control processors \citep{friedmann13phd}, which we refer to as \gls{ppu}. The overarching goal of this part of the system is to complement the flexible and configurable neuron and synapse architecture with an equally flexible digital control architecture. Here we highlight several use cases of this design choice: implementation of programmable hybrid plasticity, automatic on-chip calibration, parallel readout of analog observables for in-the-loop learning, orchestration of analog artificial neural network computation, and simulation of virtual environments.

Apart from speed, a big problem of physical implementations is their limited flexibility, especially regarding learning rules.
BrainScaleS-2 uses a ``hybrid plasticity'' scheme \citep{friedmann2016hybridlearning}, combining analog measurements with digital calculations to increase the flexibility while keeping the advantages of an accelerated physical model, like simultaneously observing all correlations between pre-and postsynaptic signals.
In a moderate-sized network, this amounts to tens of thousands of measurements per second for each synapse.
Thus, a single BrainScaleS ASIC can perform several tera-correlation measurements per second.
Compared to biological model dynamics, the emulation runs one thousand times faster;
a speedup factor that, even for small to medium-sized plastic networks, is typically out of reach to software simulations \citep{zenke2014auryn}.

To implement the plasticity rules themselves, i.e., to calculate new weights, topology information, and neuron parameters during the continuous-time operation of the network, the hybrid plasticity architecture relies on specialized build-in \gls{simd} units in the two microprocessor cores. They interface directly with the synaptic crossbar and the neuron circuits via the \gls{cadc}, which, as the name suggests, can perform simultaneous measurements of analog quantities in one row and across all \num{256} columns of the synaptic crossbar. The result of the correlation measurements can then enter plasticity programs, which can perform both fixed-point and integer arithmetic operations on vectors of either \num{128}\texttimes{}\SI{8}{\bit} or \num{64}\texttimes{}\SI{16}{\bit} entries. The weights and addresses stored in the synaptic array and the voltage traces from the neuron circuits allow for flexible plasticity computations bridging multiple timescales. In particular, we have demonstrated several versions of \gls{stdp}-based learning rules (R-STDP \citep{wunderlich2019advantages}, homeostatic plasticity \citep{cramer2020control}), as well as learning rules that compute updates based on small artificial neural networks \citep{bohnstingl2019neuromorphic} and structural plasticity \citep{billaudelle2021structural}.

The scalar part of the processor core can operate independently and take responsibility for scheduling the data-parallel instructions, data transfers, and measurements. Scoreboards track data dependencies in an in-order issue out-of-order retire scheme, which allows both the vector unit and the scalar unit to perform, for instance, arithmetic operations, that are independent of the completion of potentially higher latency load/store operations or \gls{cadc} measurements, thereby reducing overall execution time. This asynchronous operation of vector and scalar units is beneficial, in particular, for plasticity programs. 

Besides plasticity programs, the parallel access to the analog synapse and neuron state is also helpful in the gradient-based learning approaches discussed below. The surrogate gradient-based learning approach relies on \gls{cadc} samples of the membrane voltage during experiment execution. A program running on the \gls{ppu} performs this sampling in a tight loop and then writes the resulting traces to either internal SRAM or external memory. The external bandwidth of the system and the DDR3-memory the \gls{fpga} interfaces with make this approach feasible. During artificial neural network in-the-loop training and inference, the processor cores perform data transfer, accumulation of partial results, and the analog readout of matrix-vector multiplication results.

As will be discussed in the following \cref{ssec:faithful_emulation}, the neuron circuits can be adjusted by calibration of \num{24} parameters each. In addition, the parallel access to the neuron circuit membrane dynamics via the \gls{cadc} allows us to implement efficient on-chip calibration, as done in several of the experiments reported here.

Finally, the processor cores can simulate virtual environments. This use case is enabled because the processor cores can inject spikes into the synaptic crossbar and readout rates and voltage traces from the neuron circuits. It is, therefore, possible to close the agent-environment loop by simulating the agent's state and the environment interaction on the \gls{ppu} and implement the agent's action selection (in part) as a spiking neural network. We will give a detailed description of one such use case in \cref{ssec:insect_inspired_navigation}.

The flexibility of the digital architecture is in large part also enabled by a compiler, the C++ programming language, and library support. The scalar processor cores implement a subset of the \SI{32}{\bit} POWER instruction set architecture \citep{powerisa_206}. The SIMD vector instructions are custom but generally follow the conventions of the VMX SIMD instruction extension \citep{powerisa_206}. We have modified the GCC compiler toolchain to support the custom vector instructions based on the pre-existing support of the POWER instruction set architecture. With this compiler support as a foundation, we have implemented support for the C++ standard library and ensured that our hardware abstraction library (HAL) is usable both on the embedded processor cores and on the host system, with a user-transparent change of hardware access modes. Details were reported by \citet{mueller2020bss2ll}. Finally, we also provide abstractions for scheduling and executing plasticity rules.

While the on-chip memory resources are limited to \SI{16}{\kibi\byte} SRAM per processor core, the cores have access to an external memory interface, which can back both the \SI{4}{\kibi\byte} per core instruction cache and provide higher-latency access to \gls{fpga} attached memory or block RAM. This memory architecture allows for both the execution of latency-critical code without external memory access and less latency sensitive but memory intensive setup and calibration code. Future revisions will include on-chip memory controllers for direct access to suitable external memory technology.
 \section{Applications of the BrainScaleS-2 system}
\subsection{Faithful Emulation of Complex Neuron Dynamics}
\label{ssec:faithful_emulation}
\begin{figure*}[t]
    \centering
    \includegraphics{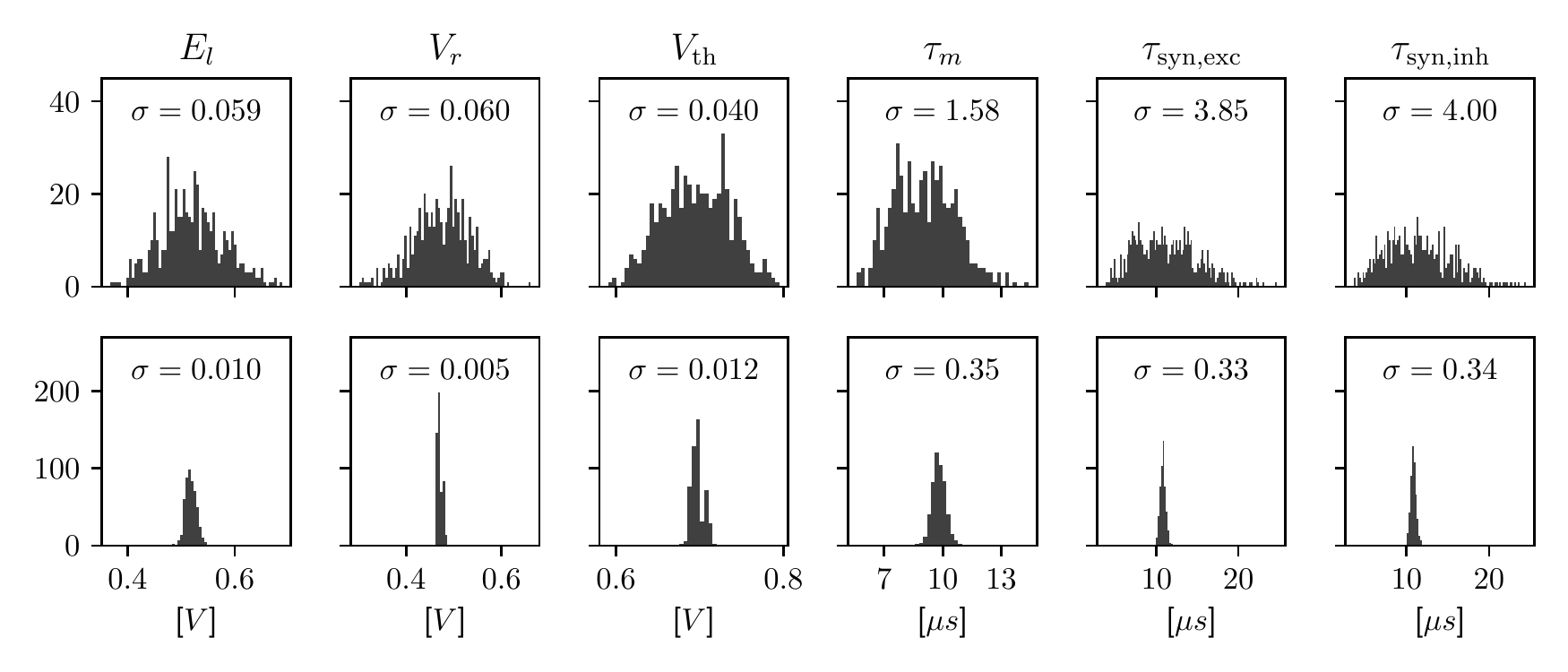}
    \caption{
        Histograms showing a characterization of \gls{lif} properties of neurons before (top) and after (bottom) calibration.
        Each histogram shows all 512 neuron circuits on a single \gls{asic}.
        In the top row, the configuration has been set equal for all neurons, to the median of the calibrated parameters.
        This results in different model characteristics due to device-specific fixed-pattern noise arising during the manufacturing process.
        After calibration, the analog parameters, such as bias currents and voltages, are selected such that the observed characteristics match a target.
    }
    \label{fig:calib-evaluation}
\end{figure*}

\begin{figure*}[t]
    \begin{center}
    	\includegraphics[width=\textwidth]{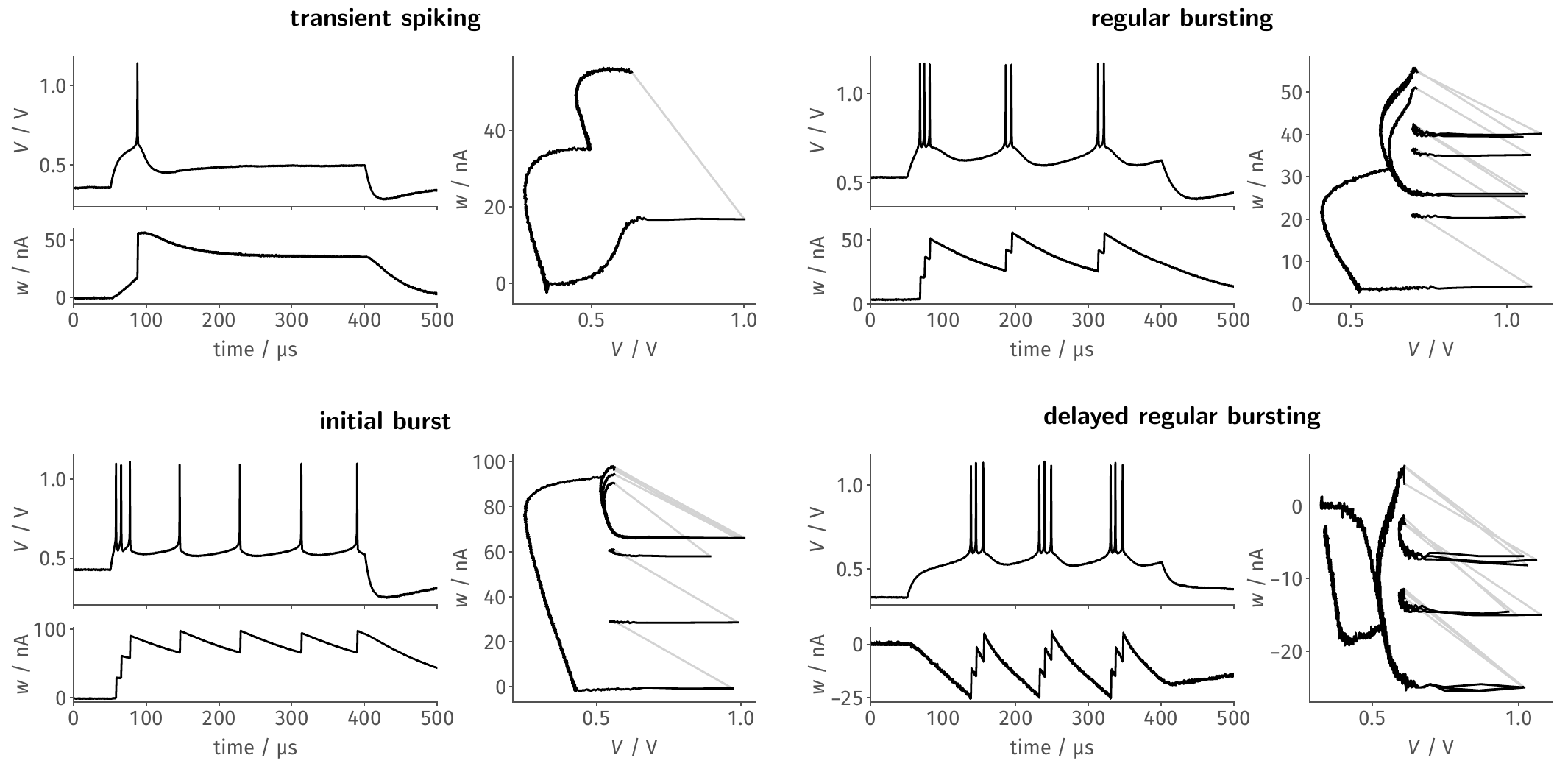}
    \end{center}
    \caption{
	    Faithfully emulating the original \gls{adex} equations, BrainScaleS-2's neuron circuits can be configured to generate distinct firing patterns as a response to a constant current stimulus.
	    Here, we tuned the neuron circuits exemplarily to replicate four of the patterns described by \citet{naud08}	using automated calibration routines.
	    Each of the four panels features the time evolution of the membrane trace and the adaptation current, as well as the resulting trajectory through the phase space.
    }
    \label{fig:firing-patterns}
\end{figure*}

Analog neuromorphic systems can usually suffer from temporal noise, fixed-pattern parameter deviations, and a divergence from the original model equations.
As elaborated in \cref{ssec:accelerated_emulation}, BrainScaleS-2 goes a long way to accomplish an extensive and at the same time detailed control over each individual circuit and thus model parameter.

We employ calibration to find configuration parameters for each circuit such that its observable characteristics match a given target.
Due to the device-specific nature of fixed-pattern deviations, this calibration is an iterative process involving a measurement on each of the specific circuits.
These measurements are a minimal hardware experiment, typically based around an ADC or spike rate measurement, in order to characterize one observable at a time.
The effects of calibration are visualized for \gls{lif} neurons in \cref{fig:calib-evaluation}.
However, the scope of calibration extends beyond those, to, e.\,g., the \gls{adex} model, multicompartment functionality, and technical parameters that don't correspond to a term in a model, but are necessary for the circuitry to behave as expected.

In this section, we want to use this high configurability to replicate the original firing patterns analyzed by \citet{naud08} and to show how the presented system can be used to emulate multi-compartmental neuron models.
For these experiments, system configuration as well as stimuli data are generated on a host computer and then transferred on the \gls{fpga}.
The \gls{fpga} handles experiment control and buffers recorded voltages.
These voltage recordings are performed with the fast on-chip \gls{adc} which offers a resolution of \SI{10}{bit} and a sampling frequency of about \SI{29}{\mega\hertz}.
At the end of the experiment, the voltage recordings are transferred to the host computer and evaluated.

\subsubsection{Replication of Biological Firing Patterns}

In order to test the full \gls{adex} model as depicted above we selected sets of model parameters from \citet{naud08} and mapped all voltages, currents, and conductances to the circuit's native domain.
Specifically, we honored the acceleration factor of \num{1000}, the physical membrane capacitance, and the voltage range of the silicon neuron.
We then tuned our neurons to these model parameters by utilizing automated calibration routines.

For each of the designated firing patterns, we then stimulated the neurons with a respective step current $I_\text{stim} = I_0 \cdot \Theta(t - \SI{50}{\micro\second}) \cdot \Theta(\SI{400}{\micro\second} - t)$.
We recorded the membrane potential as well as the adaptation state voltage and -- based on the latter -- estimated the adaptation current flowing onto the membrane.
\Cref{fig:firing-patterns} shows these two resulting traces as well as the resulting trajectory through phase space for a single neuron.
The four exemplarily chosen firing patterns highlight different aspects of the neuron design:
\emph{Transient spiking} requires both large spike-triggered adaptation increments $b$ and a strong subthreshold adaptation $a$ to emit a single spike as a response to the stimulus onset and remain silent for the remainder of the current pulse.
\emph{Regular bursting} and the \emph{initial burst}, in contrast, already emerge for configurations with small $a$ and mainly rely on spike-triggered adaptation.
These two patterns primarily differentiate themselves in the exact choice of the reset potential $V_\text{r}$ in relation to the threshold voltage $V_\text{T}$, which demonstrates the precise control over the respective circuit parameters.
Finally, \emph{delayed regular bursting} relies on an inverted subthreshold adaptation ($a < 0$) leading to positive feedback.

For the last three patterns, the trajectory through the phase plane spanned by $V$ and $w$ nicely demonstrates the precise and reproducible dynamics of the circuit, which are especially highlighted by the remarkable stability of the limit cycles for the periodic spiking activity during stimulation.
While here only shown for a single neuron, all patterns could -- without manual intervention -- be reliably reproduced by most, if not all, of the neuron circuits.
For example, regular bursting could be simultaneously configured for all of the \num{128} tested neurons.

\subsubsection{Multi-compartmental Neuron Models}

\begin{figure}
    \includegraphics[width=0.5\textwidth]{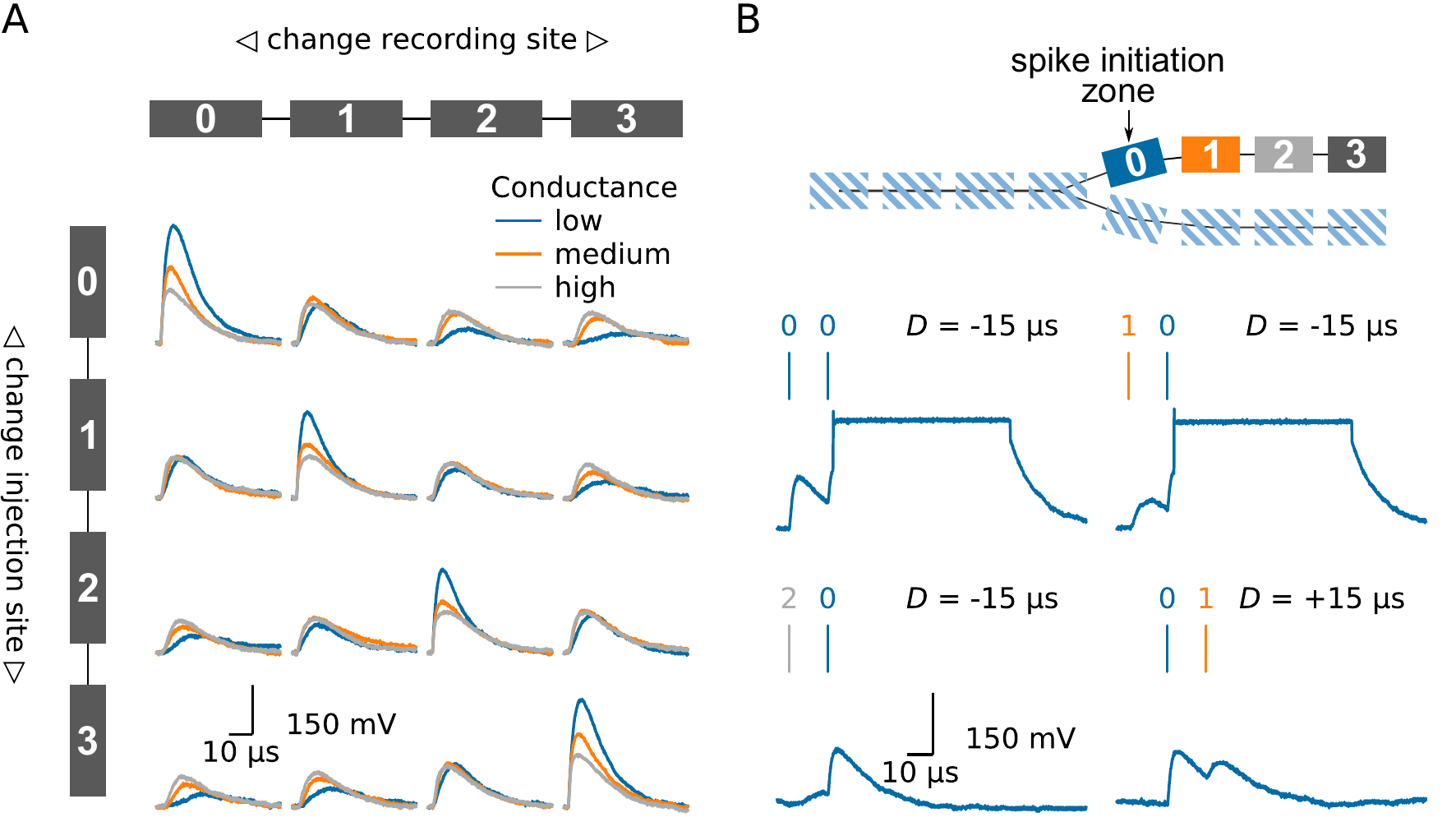}
    \caption{Emulating multi-compartmental neuron models on BrainScaleS-2.
            (A) Four compartments are connected to form a chain, compare model at the top and left side.
            The traces show the membrane potential in the four different compartments as a synaptic input is injected in one compartment after another.
            Thanks to the high configurability of the BrainScaleS-2 system, compare \cref{ssec:accelerated_emulation}, the conductance between the individual compartments can be altered to control the attenuation of the signal along the chain.
            (B) Model of a chain which splits in two.
            The BrainScaleS-2 system supports dendritic spikes, here demonstrated in the form of plateau potentials.
            Inputs are injected in different compartments with a fixed delay between them; this is indicated by the vertical bars above the membrane traces.
            Depending on the spatio-temporal distribution of the inputs a dendritic spike can be elicited.
            The traces show the membrane potential in \emph{compartment 0}.
            Figure adapted from \citet{kaiser2021emulating}.}
    \label{fig:compartment-chain}
\end{figure}

To further demonstrate the high configurability of the system at hand we want to show how multi-compartmental models can be emulated.
As mentioned in \cref{ssec:accelerated_emulation}, BrainScales-2 offers the possibility to connect several neuron circuits and therefore allows to implement various compartmental neuron models \citep{kaiser2021emulating}.

A passive compartment chain model can for example be used to replicate the behavior of a passive dendrite, \cref{fig:compartment-chain} A.
We inject synaptic input in one of the compartments and investigate how the \gls{epsp} travels along the chain of compartments.
For that purpose, we once again use the fast on-chip \gls{adc} to record the membrane potentials in the different compartments.
In the first row the input is injected into the left compartment.
As expected, the \gls{epsp} becomes smaller and broader as it travels along the chain.
The extent of the attenuation can be controlled by the conductance between the different compartments.
When looking at the case where the input is injected into the second compartment the influence of the neuron morphology becomes obvious.
The height of the \gls{epsp} is smaller as compared to the injection in the first compartment.
This is due to the lower input conductance: the second compartment has two neighbors as compared to the single neighbor for compartments at the end of the chain.

Dendrites are not simple passive cables but are able to initiate local regenerative events \citep{schiller2000nmda, larkum1999cellular, major2013active}.
On BrainScaleS-2 each compartment is made up of one or more fully functional neuron circuits and can therefore replicate dendritic spikes.
Besides sodium-likes spikes, which are modeled by the AdEx model, the neuron also supports plateau-like spikes.
\Cref{fig:compartment-chain} B illustrates the model of a dendritic branch which splits into two thinner dendrites.
The first compartment, \emph{compartment 0}, of one of the branches is configured to initiate plateau potentials.
We inject synaptic inputs in two compartments with a fixed delay between them and record the membrane potential at the spike initiation zone.
As expected, spiking depends on the spatio-temporal distribution of the inputs \citep{williams2002dependence, polsky2004computational}.
While inputs near the spike initiation zone elicit a dendritic spike, more distal inputs fail to cause a threshold crossing.
Furthermore, a spike is more easily triggered if the distal input precedes the input at the initiation zone.
 \subsection{Biology-Inspired Learning Approaches}
\label{ssec:biology_inspired_learning_approaches}
One underlying goal of the system is to enable the exploration of biologically plausible learning rules at accelerated time scales relative to biology. For our purposes, we consider a learning rule or algorithm to be biologically plausible if it satisfies several criteria. The algorithm should be \emph{spatially} and \emph{temporally} local. By spatially local, we mean that parameter changes computed by the algorithm should rely only on observations that can be locally made at each neuron and synapse. Some aspects of biological plausibility are enforced by the system design itself. The \glspl{ppu} -- coupled to the correlation circuitry implemented in each synapse -- facilitate the implementation of such spatially local algorithms, as we have discussed in \cref{ssec:hybrid_plasticity_and_versatile_digital_control}. A temporally local algorithm should not rely on complete traces of activity but sparse temporal observations.

Here we highlight experiments that have effectively used the underlying hardware capabilities to demonstrate aspects of biologically plausible learning. One aspect of biological learning systems is that they typically interact with an environment. We have realized tasks on the BrainScaleS-2 system in which an agent interacts both with a simulated and physical environment. Tasks range from simple Markov decision processes like maze navigation and Multi-Armed Bandits \citep{bohnstingl2019neuromorphic}, playing a version of Pong \citep{wunderlich2019advantages}, to insect navigation and control of an accelerated robot \citep{schreiber2021accelerated}. In all of these instances, the rapid reconfigurability and experiment execution time lead to a significant speedup over a simulation on commodity hardware.

The virtual environment and the agents are simulated on the plasticity processor, which can guarantee low latency due to its tight coupling to the analog neuromorphic core. In fact, this lack of deterministic low latency coupling made such experiments difficult on the BrainScaleS-1 systems, where that problem was further emphasized by the additionally increased acceleration factor.

Besides the three reinforcement learning related experiments, we also explored other aspects of biologically plausible learning. For example, the parallel access of the plasticity processing unit to both the (anti-)correlation sensor readings and the digital weight and address settings of the synapse array suggests experiments based on synaptic rewiring and pruning \citep{billaudelle2021structural}. This work makes use of the ability of our synaptic crossbar to realize sparse connectivity, as each synapse has a local receptive field of \num{64} potential inputs. Last but not least, we have performed work exploring criticality and collective dynamics of spiking neurons subject to homeostatic plasticity rules on both the scaled-down prototype systems \citep{cramer2020control} and in ongoing work, which we will report on in \cref{ssec:collective_dynamics}.

\subsubsection{Insect-Inspired Navigation}
\label{ssec:insect_inspired_navigation}
One way to study neural computation is to focus on small functional circuits. While the scale of the single-core system with its \num{512} neuron circuits and the prototype systems with \num{32} neurons do not allow the exploration of large-scale dynamics, they should enable the study of functional circuits over long periods. Here we focus on a recently published anatomically constrained model of the path integration abilities of the bee brain \citep{stone2017anatomically}. Path integration enables bees to return to their nest after foraging for food successfully. In adopting the model to the hardware constraints, we had to both translate the rate-based model to a spike-based model and find a way to implement the integration primitives in terms of the plasticity mechanisms available in hardware. The resulting model is scaled-down relative to the model proposed by \citet{stone2017anatomically} to fit the hardware constraints of the prototype system. Furthermore, the model can be run at the \num{1000}\texttimes{} accelerated time scale relative to biology on a prototype of the BrainScaleS-2 system and, more recently, on the full-scale single-chip system. 

The model is evaluated on a task divided into three parts. In the first ``foraging'' part, the bee performs a random walk starting at the ``nest'' location, with two sensory neurons receiving light direction information. In a second ``return'' phase output of the motor neurons is used to determine the movement of the bee back to the nest. Finally, during a ``looping'' phase, the bee is supposed to remain as close as possible to the nest until the experiment is terminated.

\begin{figure*}[ht!]
\includegraphics[width=\textwidth]{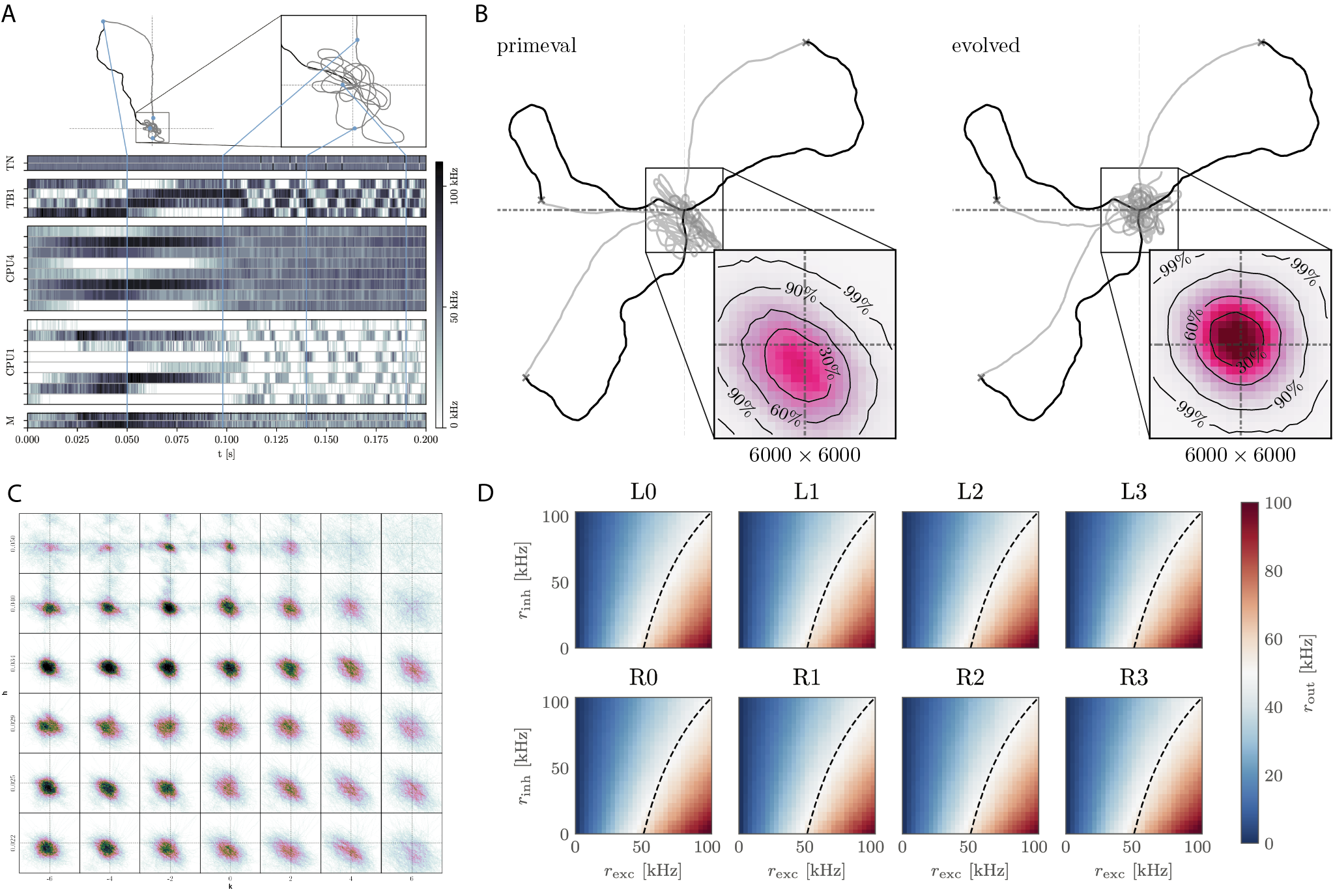}
\caption{Insect-inspired path integration and navigation on the BrainScaleS-2 accelerated neuromorphic substrate. (A) Network activity and path during a single experiment instance. The simulated agent first randomly spreads out from the ``nest'' until it finds ``food''. It then returns to the nest using the integrated path information. Once returned, it starts looping around the nest until the experiment is terminated. The blue lines indicate how points along the trajectory correspond to the activity trace shown below. From top to bottom, the neuron activity traces are that of the two sensory neurons (TN), ``compass'' neurons (TB), integrators (CPU4), steering (CPU1), and motor neurons (M). Signals flow from $\text{TN} \to \text{TB}$, $\text{TB} \to \text{CPU4, CPU1}$; $\text{CPU4} \to \text{CP1}$ and $\text{CP1} \to \text{M}$. Details of the network architecture were given by \citet{schreiber2021accelerated}. (B) We performed evolutionary optimization of the weights controlling the behavior of our agents. Here
we show \num{3} sample trajectories each of the initial (primeval) and evolved population. Inset is a histogram over \num{1000}
trajectories in the final ``looping'' phase of the evaluation, zoomed in to a \num{6000}\,\texttimes\,\num{6000} square of positions at the origin. As can be seen, the evolved population reaches a more symmetrical and tighter looping behavior. (C) Influence of two hyperparameters $h$, $k$ on the integration performance of the spiking integrators, each \num{12000}\,\texttimes\,\num{12000} square of positions contains a histogram of \num{100} trajectories in the looping phase $(t > 2 t_{\text{return}})$ for a total of \num{4200} trajectories. (D) Response to excitatory and inhibitory input rates of the calibrated CPU1 neurons. The dashed line indicates where \SI{50}{\percent} of the maximum output rate is expected. Calibration of these neurons, as well as all the other neurons, was done \emph{on-chip} using the \gls{ppu}, in the case of the implementation done on the full-scale BrainScaleS-2 system. Panels (A,B,C,D) are adapted from \citet{schreiber2021accelerated}.}
\label{fig:insect_navigation}
\end{figure*}

Details of the signal flow and an activity trace are given in \cref{fig:insect_navigation} A. A ``foraging'', ``return'' and ``looping'' episode was simulated in \SI{200}{\milli\second}, therefore corresponds to \SI{200}{\second} in biological time. The virtual environment of the bee, as well as simulating the bee's ``foraging'' random walk, position, heading direction in the environment, and sensory perception, are implemented on the \gls{ppu}. It, therefore, closes the perception-action loop in conjunction with the accelerated emulation of the path integration circuit on the neuromorphic core. Moreover, the behavior of the CP4 integrator neurons relies on weight modifications of axoaxonic synapses, which are implementable due to the flexible synaptic modifications that can be performed by the \gls{ppu}. 

The acceleration factor also allows us to efficiently sweep the hyper-parameters of the integrator neurons (\cref{fig:insect_navigation} C). Finally, we performed an evolutionary optimization in order to improve the agents' distance to the nest during the looping phase (\cref{fig:insect_navigation} B). Without the acceleration factor, the hyperparameter sweep would have required \SI{9.6}{\day}. Instead, it can be performed in approximately \SI{14}{\minute}. Similarly the evolutionary optimisation over \num{200} epochs of \num{1000} individuals would have required \SI{448}{\day}, instead it could be completed in roughly \SI{12}{\hour}. The output data required by the evolutionary algorithm is buffered by the FPGA and evaluated on the host. The host is also responsible for the initial and all subsequent configurations of the system, as well as the implementation of the evolutionary algorithm.

Therefore, we believe this to be a compelling case study of the system's modeling capabilities concerning small embodied neuromorphic agents and of how contemporary research in computational neuroscience could benefit from an accelerated physical emulation platform to evaluate experiments. While implementing such a model on neuromorphic hardware enforces constraints on the modeler not present in a free-form von-Neumann modeling approach, it also has benefits. On a technical side, the acceleration factor is guaranteed even when scaling to larger circuits, including plastic synapses. On the conceptual side, many of the constraints present in the hardware, namely spike-based communication and local plasticity rules, are also believed to be present in biology. Therefore, one is encouraged to fully commit to a modeling approach without any shortcuts afforded by using a von-Neumann architecture to simulate the model. Furthermore the experiment demonstrates that it is possible to calibrate the neuron circuits with on-chip resources, which is crucial for a scalable and host-independent calibration.

A detailed description of the implemented spike-based model and the experiments, including the evolutionary optimization, was given by \citet{schreiber2021accelerated} and will be further elaborated on in a forthcoming publication \citep{schreiber2022insectoidpath}.

\subsubsection{Accelerated Closed Loop Robotics}
\label{ssec:accelerated_robotics}
\begin{figure*}[ht!]
\includegraphics[width=\textwidth]{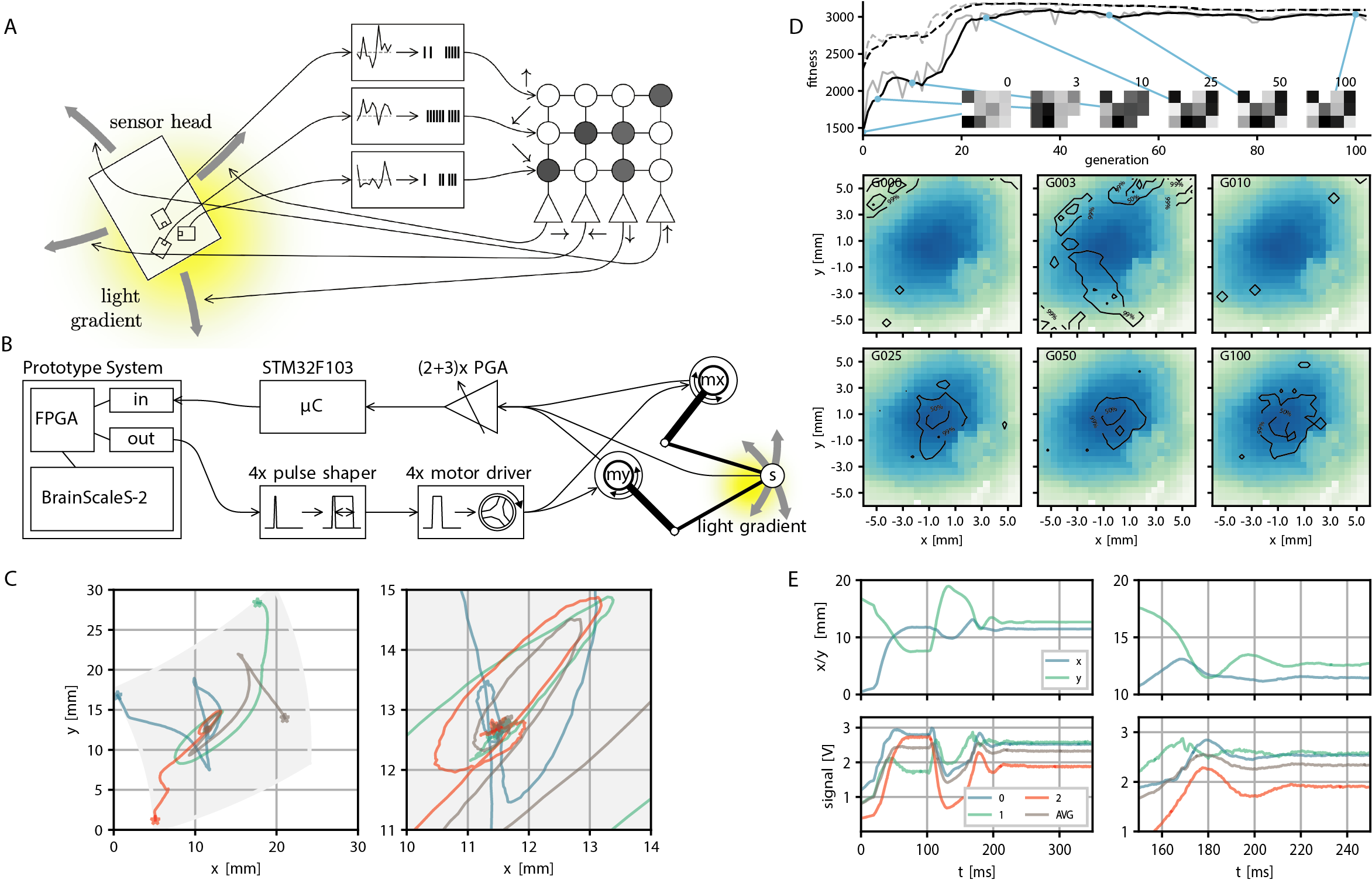}
	\caption{Real world closed loop interaction with the BrainScaleS-2 system. (A) Conceptual overview of the example experiment, a sensor head with three light sensors is moved over an illuminated surface. Sensor readings are converted into spike input to \num{3} input sources in a small feedforward spiking neural network. Spike output produced by \num{4} neurons is converted into motor commands, which move the sensor head on the surface. The goal is to follow the light gradient. (B) Physical realisation of the example experiment from right to left: The PlayPen2 consists of two actuated arms which pantographically move a sensor head over a screen or illuminated surface. Signals from the sensor head are digitally processed by a micro controller and converted into spikes send into a \gls{fpga} used to interface with a scaled down BrainScaleS-2 prototype system, which implements the small feedforward spiking network. Spike outputs are routed in the \gls{fpga} to the the Spike I/O interface and converted by pulse shapers into motor command pulses. (C) Example trajectories of the sensor head on the surface, with grey indicating the accessible region (left) and zoom in on the center region where the brightness maximum is (right). (E) Position (top) and brightness signals (bottom) of the sensor head over time. The two panels on the left show the full time course. The neural network starts control at $t = \SI{100}{\milli\second}$ and stops at $t = \SI{250}{\milli\second}$. The two panels on the right show a zoom in on the interval $t \in [100,200] \text{ms}$, with the grey curve indicating an average over all brightness readings. (D) We perform evolutionary optimisation of the \num{4}\,\texttimes\,\num{3} weight matrix both from a random initial weight configuration. We show the moving average of the fitness in black and the fitness at a certain generation in grey, both for the top three individuals (dashed line) and the population average (solid line). In addition we display the weight configuration of an arbitrary individual at \num{6} selected generations (\numlist{0;3;10;25;50;100}). Finally we show a qualitative performance evaluation at the same  generations of the average weight matrix over \num{100} experimental runs divided into \num{4} starting positions. The contourlines show where \SI{99}{\percent}, \SI{50}{\percent} and \SI{10}{\percent} of all trajectories ended after \SI{225}{\milli\second}, when the maximum brightness should be reached. The contourlines are overlayed over a plot of the brightness sampled from the photo diode signals over all runs and generations. Missing pixels correspond to locations not reached at any time. Panels (A-E) were adapted from \citet{schreiber2021accelerated} and (C) from \citet{billaudelle2020versatile}.}
\label{fig:accelerated_robotics}
\end{figure*}

We can also consider another family of applications: Using the BrainScaleS-2 system to control another physical system. This presents several challenges: The accelerated neuron dynamics means that the natural time scale at which the system could interact with a real environment would be on the order of microseconds. Similarly, sensor information to be processed by the system needs to be on this time scale, at least if spikes are meant to be used as the information processing primitives. As a case study, \citet{schreiber2021accelerated} implemented an accelerated mechanical system consisting of two actuators moving a light sensor over an illuminated surface or screen. Here we give a brief overview and refer to \citet{schreiber2021accelerated} for a detailed exposition. While this robotic design has limited practical purpose, it illustrates the key challenges any such attempt faces. This begins with the challenge of translating signals between the robot and spiking domain, where the challenge is to interpret the sensor reading as spikes naturally and to convert the ``motor'' neuron output into actuator input. Here, we solve this in two different ways: In a first prototype, the sensor to spike conversion was done by connecting modular analog circuits, featuring circuits for differentiation, inversion, adder, noise, and spike generation, which ultimately could produce spike input to a parallel \gls{fpga} interface. While this came closest to the ideal of analog and spike-based communication, we replaced this component with a fully digital micro-controller-based solution in a second iteration. This implementation choice is beneficial because other experiments could also use this microcontroller to implement more sophisticated virtual environments. In both versions of the design, spikes of the motor neurons with a width of roughly \SI{500}{\nano\second} were converted into actuator input by using pulse-shapers. This required motor drivers and actuators, which could meaningfully react to input pulses with approximately \num{10} microsecond duration. Fortunately, voice coil actuators, as used in commodity hard drives, precisely have this property. \Cref{fig:accelerated_robotics} B, shows an overview of the signal path. The actuators move a sensor over the illuminated surface or screen. In \cref{fig:accelerated_robotics} we show task performance and sensor trajectories of a ``maximum finder'', that is, the task of the agent is to find the local intensity maximum following the light intensity gradient. And indeed, the following of (chemical) gradients is one way in which biological organisms find food sources. The accelerated nature of the experiment execution allows for fast evolutionary optimization of the required weight matrix (see \cref{fig:accelerated_robotics} D, E). Other experiments demonstrate that, for example, one can construct networks capable of rapidly detecting light intensity ``edges''.

Beyond the experiments done so far, it would also be possible to display a maze environment on the screen and let the neuromorphic agent physically solve the maze, with rewards being encoded either virtually by the micro-controller or by light intensity changes in the maze. Although initial calculations suggested that the speed would be sufficient to produce light intensity changes at the right time scale, ultimately, we were only able to achieve regulation oscillations with a period of approximately \SI{10}{\milli\second}. While this is still quite fast, it does not naturally correspond to the time scale of the system. In some sense, the neuromorphic system, therefore, can only control this environment in slow motion. This is less of an issue for a maze-like environment, where the decision procedure could be more abstractly interpreted in terms of ``moves''.

Besides the initial configuration and the implementation of the evolutionary optimisation, which requires recorded data buffered on the FPGA and transferred to the host, the experiment is implemented in a host independent fashion. It therefore demonstrates accelerated closed-loop neuromorphic computation. 

Overall we believe using our accelerated neuromorphic system for rapid spike-based control and sensing has many exciting future applications. Further experiments can extend the proof of principle we present here in several directions. For example, instead of voice coils, one could control piezo-electrically actuated mirrors or lenses to divert laser light in a physical experiment or control micro-sized or high-speed aerial vehicles. Other high-speed applications could include motor control circuits and ultra-sound or radar applications, particularly involving active sensing or phased arrays.

\subsubsection{Collective Dynamics}
\label{ssec:collective_dynamics}
\begin{figure*}[ht!]
    \centering
    \includegraphics{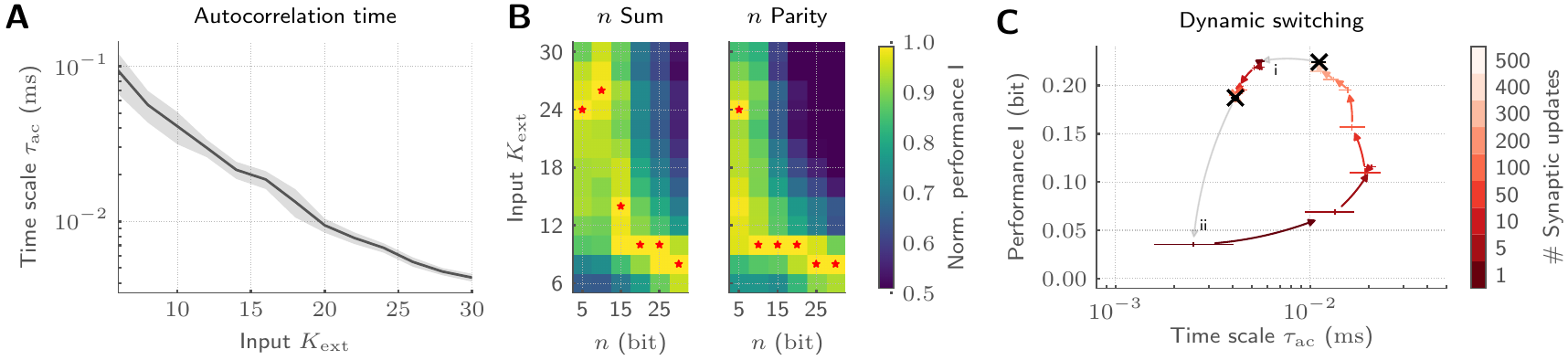}
    \caption{Exploiting collective dynamics for information processing on BrainScaleS-2.
	    (A) The autocorrelation time $\tau_\mathrm{ac}$ of recurrent SNNs can be controlled by changing the input strength $K_\mathrm{ext}$ under the constant application of homeostatic regulation.
	    (B) The emerging autocorrelation times for low $K_\mathrm{ext}$ can be exploited for complex, memory-intensive task processing.
	    Both, a $n$ bit sum as well as a $n$ bit parity task profit from the complex processing capabilities for high $n$.
	    Low $n$ task, in contrast, profit from short time scales and hence high $K_\mathrm{ext}$.
	    The optimal $K_\mathrm{ext}$ for each $n$ is highlighted by red stars.
	    As a result, each task requires its own dynamic state.
	    (C) The adaptation of the network dynamics to task requirements can be achieved by switching the input strength under the constant action of homeostatic regulation irrespective of the initial condition.
	    Here, the transition from a state with long time scales to short ones is completed with only a few homeostatic weight updates (i).
	    The reverse transition requires a longer relaxation phase (ii).
	    }
    \label{fig:collective_dynamics}
\end{figure*}

One way to adapt recurrent \glspl{snn} to perform information processing is to deliberately exploit collective dynamics.
Particularly promising are the dynamics emerging at a so-called \emph{critical point} at which systems fundamentally change their overall characteristics, transitioning between e.\,g.\, order and chaos or stability and instability.
Being at this point, systems maximize a set of computational properties like sensitivity, dynamic range, correlation length, information transfer and susceptibility \citep{harris2002,munoz2017,barnett2013,tkavcik2015}.

Here, we showcase the tuning of plastic recurrent \glspl{snn} to and away from criticality by adapting the input strength $K_\mathrm{ext}$ on a prototype of the BrainScaleS-2 system \citep{cramer2020control}.
The \gls{cadc} as well as the \gls{ppu} facilitate an on-chip implementation of the \gls{stdp}-based synaptic plasticity required to adapt the collective dynamics of our \glspl{snn}.
Within these experiments, the \gls{fpga} is only used for experiment control as well as spike injection.
On the latest chip revision, the latter can be achieved by drawing on the on-chip spike generators of BrainScaleS-2, thereby reducing the strain on I/O.
Here, the hybrid plasticity approach in combination with the accelerated nature of the BrainScaleS-2 architecture allows us to fully exploit the associated advantages by bridging the gap in time scales between neuro-synaptic dynamics, network dynamics, plasticity evaluation as well as acquisition of long-lasting experiments for statistical analysis.

With our implementation, we showcase emergent autocorrelation times for low input strengths $K_\mathrm{ext}$, significantly exceeding the time scales of single-neuron dynamics (\cref{fig:collective_dynamics}A).
Most notably, adjusting $K_\mathrm{ext}$ allows us to precisely tune the time scale of the collective dynamics.
These dynamics can be deliberately exploited for information processing by tying on the reservoir computing framework \citep{jaeger01echo,maass02realtime}.
To that end, we characterize the interplay of collective dynamics and task complexity by training a linear classifier on the host computer based on the spike trains emitted by the BrainScaleS-2 chip.
While long time scales for low $K_\mathrm{ext}$ only boost the performance for complex memory-intensive tasks, simple tasks profit from short intrinsic time scales (\cref{fig:collective_dynamics}B).
Hence, every task requires its own dynamics to be solved optimally \citep{cramer2020control}.
The required tuning can be realized by dynamically adjusting $K_\mathrm{ext}$ under the constant action of the synaptic plasticity.
This switching can again be efficiently achieved on BrainScaleS-2 by exploiting the acceleration.
The transition from a state with high to low autocorrelation time is completed with \num{50} synaptic updates amounting to only \SI{50}{\milli\second}, whereas the reverse transition requires \num{500} updates and hence \SI{500}{\milli\second} (\cref{fig:collective_dynamics}C).
Moreover, this switching leads to comparable dynamical regimes irrespective of the initial condition and is characterized by a low energetic footprint when drawing on the on-chip spike sources.
With this, we provide not only an understanding of how the collective dynamics can be adjusted for efficient information processing, but in addition, showcase how the physical emulation on BrainScaleS-2 allows to bridge the vast range of time scales in the associated experiments which render equivalent implementations on conventional hardware prohibitively expensive.
 \subsection{Gradient-Based Learning Approaches}
\label{ssec:gradient_based_learning_approaches}
\begin{figure*}[t!]
	\includegraphics[width=\textwidth]{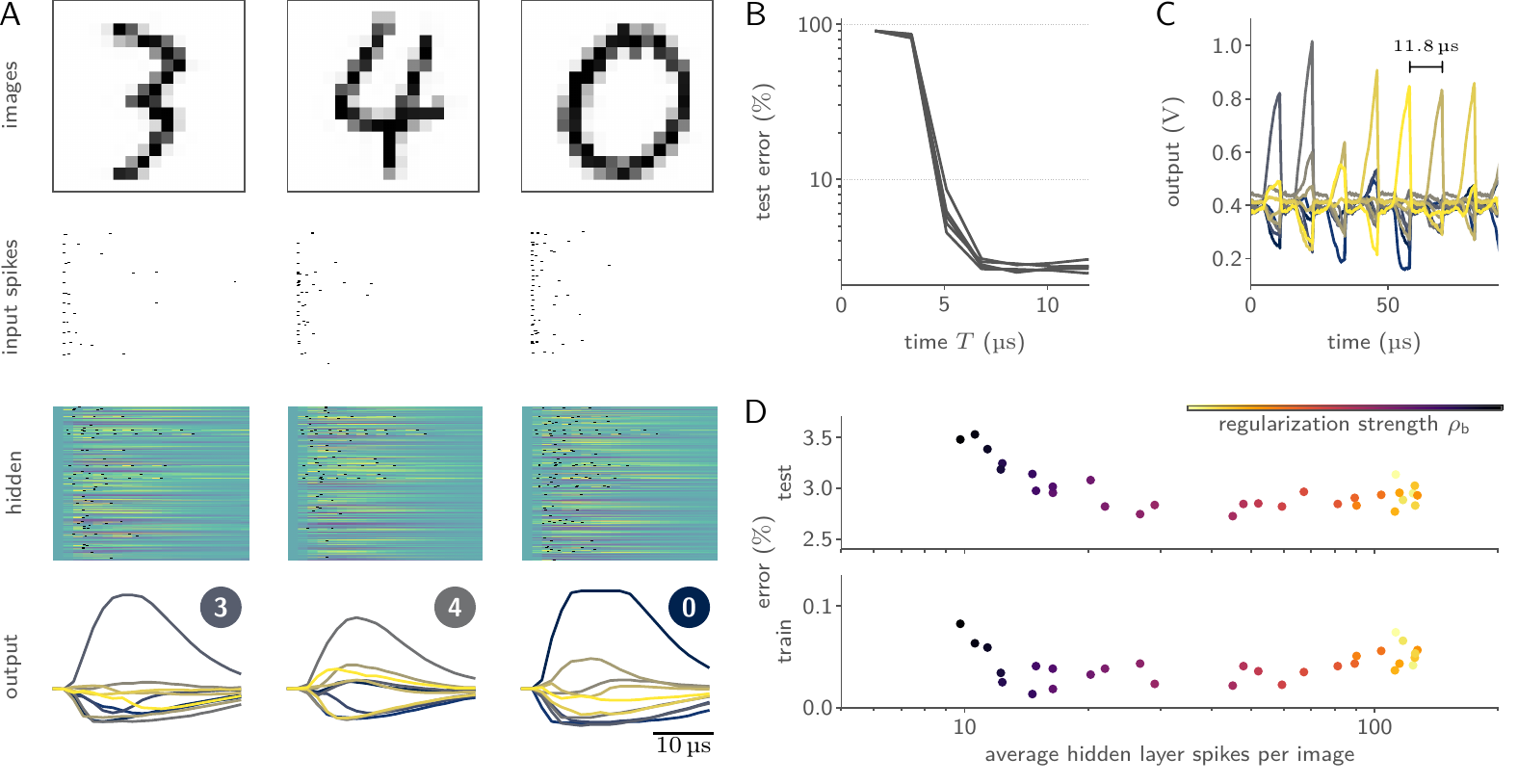}
	\caption{Training of and inference with \glspl{snn} on BrainScaleS-2 using surrogate gradients.
		(A) Exemplary activity snapshots for three \num{16}\,\texttimes\,\num{16} pixel MNIST images, the resulting spike-latency encodings, an overlay of the hidden layer membrane traces recorded at a data rate of \SI{1.2}{\giga\bit\per\second} as well as the resulting hidden layer spikes, and the output layer membrane traces.
		(B) Spike-latency encoding promotes fast decision times: after \SI{7}{\micro\second} the network reaches its peak performance.
		(C) A fast inference mode allows to exploit the quick decision times by artificially resetting the neuronal states after \SI{8}{\micro\second} and therefore preparing the network for presentation of the next sample, culminating in a classification rate of \SI{85}{\kilo\nothing} inferences per second.
		(D) The PyTorch-based framework allows to co-optimize for near-arbitrary regularization terms, including sparsity penalties. In this instance, BrainScaleS-2 can be trained to classify MNIST images with an average of \num{12} hidden layer spikes per image without a significant decline in performance.
		}
	\label{fig:strobe}
\end{figure*}

Both the individual analog circuits and the overall system can be considered to be parametrized physical systems. A particular task can be represented as a constraint optimisation problem involving a loss function and constraints which implement the input-output relation. One approach to such a constraint optimisation problem is to estimate gradients of the parameters and subsequently perform some form of (stochastic) gradient descent. Compared to digital computers the analog nature of the core components leads to additional challenges. Whereas digital neuromorphic systems mainly need to be concerned with the limited precision in their digital arithmetic and otherwise can exactly simulate the
operation of their system, this is not the case for analog neuromorphic systems. 

Just like with any other parametrized physical system it is important to have a model of its behavior in order to perform this optimisation. As a simple example to keep in mind, think of a physical pendulum, such as a ball hanging on a piece of string. The physical parameters of such  a pendulum, namely the length of the string $L$ and the mass of the ball $m$ enter any \emph{model} of this system. A good model of a physical system does not necessarily need to capture all the details of the physical situation to be useful. For example for small initial angles the motion of a pendulum is well described by a damped harmonic oscillator. Measuring the behavior of the pendulum on a set
of example trajectories then allows one to fit the model parameters to get good agreement between model and observed behavior. Since our analog neuromophic core attempts to replicate the dynamics of certain idealized neuron models, we are in a similar situation. Part of the correspondence between our physical substrate and the model is ensured by \emph{calibration}. During task specific training, we adapt a model to a specific hardware instance, by training \emph{in the loop}.

The \emph{in-the-loop} training paradigm relies on the fact, as we alluded to above, that it is possible to use the parameter gradient computed based on measurements and a model of a physical system to update the parameters of the physical system in a composable fashion. This is the basis of all three gradient-based learning paradigms realized in the BrainScaleS-2 system so far, namely
\begin{itemize}
\item Time-to-first spike \citep{goeltz2019fastNatMI}
\item Surrogate-Gradient-Based Learning \citep{cramer2020training}
\item Analog ANN training \citep{weis2020inference}
\end{itemize}
They differ in which measurements are necessary and what model of the physical system is used. In the time-to-first spike gradient-based training scheme, which we won't discuss in detail here, the essential idea is that it is possible to compute the derivative of the spike time with respect to input weights based on an analytical expression of the spike time. In other words the only measurements required of the system are the spike times of the neurons present in the (feed-forward) network. The model assumes that the physical system evolves according to dynamics with certain ratios between synaptic and membrane time-constants, which need to be ensured by calibration. In the surrogate gradient paradigm the network dynamics is modelled by a recurrent neural network (RNN) closely corresponding to the continuous time dynamic. It requires the observation of the membrane voltages of all neurons of the network with a temporal resolution comparable to that of the chosen RNN timestep. We will discuss this approach in more detail in \cref{ssec:surrogate_gradient_itl}. Finally in the analog ANN training mode the behavior of the system is modelled by a linear operation (implemented by the synaptic crossbar), together with a non-linearity (implemented by the digital processor and the analog readout). Again the actual correspondence of the system to this behavior needs to be ensured by calibration. In order to estimate gradients the layerwise results are needed and a full precision version of the implemented operation is used to propagate errors between layers. We give a more detailed description of this approach in section \cref{ssec:gradient_based_ann}.

\subsubsection{Surrogate-Gradient-Based Training of \acrshortpl{snn}}
\label{ssec:surrogate_gradient_itl}

Gradient-based training of \glspl{snn} has historically been impeded by their binary nature, and was mostly limited to rate-based codings schemes.
Surrogate-gradient-based approaches have only recently enabled the optimization of \glspl{snn} eliciting temporally sparse spiking activity \citep{neftci2019surrogate}.
To that end, these approaches attach a modified derivative to the neurons' activation functions and thereby smoothen their gradients.
Relying on the temporally resolved membrane potential, these surrogate gradients can often be easily computed for numerically evolved \glspl{snn}.
For analog systems, the neurons' membrane potentials evolve as physical quantities and are hence not directly available for the respective gradient computation; digitization is complicated by the intrinsic parallelism of such devices.
BrainScaleS-2, however, does allow for the parallel digitization of membrane traces, despite its accelerated nature.
For this purpose, we employed the massively parallel \glspl{adc} and scheduled their conversion via the on-chip \glspl{ppu}.
This allowed us to parallelly digitize the temporal evolution of the membrane potentials of \num{256} neurons with a sampling period of \SI{1.7}{\micro\second}.
Based on the digitized membrane traces and spike times, we then constructed a PyTorch computation graph based on the \gls{lif} equations.
By incorporating the actual, measured traces into this model of our system, we aligned the computation graph with the actual dynamics of our silicon neurons.
Our framework, hence, effectively attached gradients to the otherwise non-differentiable physical dynamics and allowed to minimize arbitrary loss functions via \gls{bptt} in combination with state-of-the-art optimizers \citep{kingma2014adam}.

We benchmarked our learning framework on several challenging datasets.
For example, we trained feed-forward \gls{snn} with a hidden layer composed of \num{246} \gls{lif} neurons on the handwritten MNIST digits (\citealt{lecunmnist}, \cref{fig:strobe}A), where we reached a test accuracy of \SI{97.6 \pm 0.1}{\percent} \citep{cramer2020training}.
Notably, we observed a time-to-decision of less than \SI{7}{\micro\second} (\cref{fig:strobe}B).
We exploited this low classification latency in a fast inference mode, where we artificially reset the analog neuronal states to prepare the network for the subsequent input sample, and reached a classification throughput of \SI{84}{\kilo\nothing} images per second (\cref{fig:strobe}C).
The flexibility of our framework could furthermore be demonstrated by augmenting the loss term with a sparsity penalty.
This regularization allowed us to perform inference on the MNIST data with on average only \num{12} hidden layer spikes per image (\cref{fig:strobe}D).
Moreover, our framework could be extended to facilitate the training of recurrent \glspl{snn}.
Specifically, we trained a recurrent \gls{snn} with a single hidden layer composed of \num{186} \gls{lif} neurons on the SHD dataset \citep{cramer2020heidelberg}, where we reached a test performance of \SI{80.0 \pm 1.0}{\percent} \citep{cramer2020training}.

\subsubsection{Artificial Neural Networks on BrainScaleS-2}
\label{ssec:gradient_based_ann}
\begin{figure}
    \includegraphics[width=0.5\textwidth]{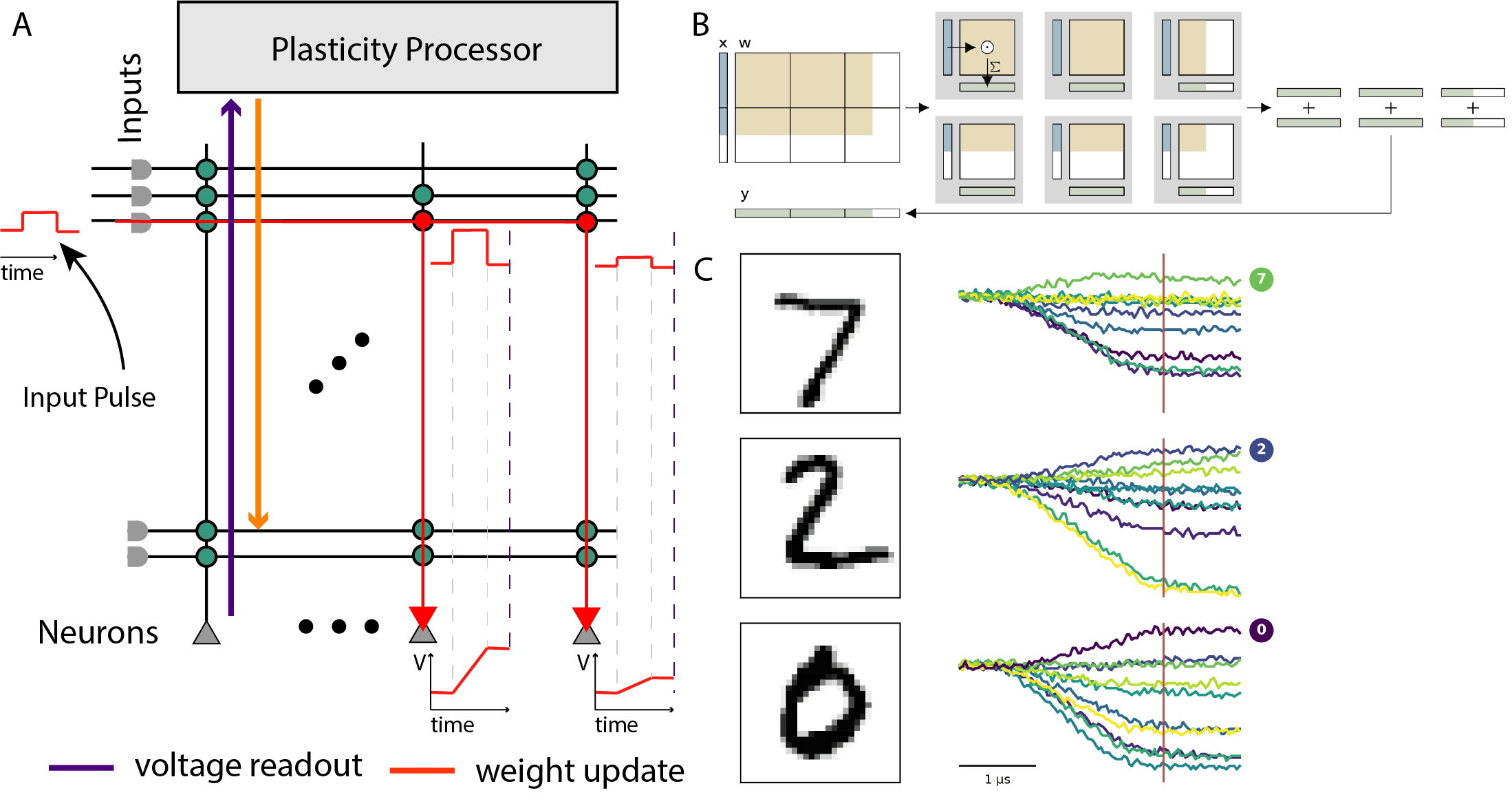}
    \caption{Use of the BrainScaleS-2 system in ANN operating mode.
        (A) Analog input values are represented as pulses sent consecutively into the synaptic crossbar. The duration of these pulses represents input activations. Synapses modulate the pulse height depending on the stored weight. Signed weights can be achieved by using two synapse rows, excitatory and inhibitory, for the same input. The neuron circuits serve as an integrator of the generated currents. Readout of the voltage occurs in parallel once triggered (dashed violet line)  by the columnar ADC.
        (B) Layers that exceed the hardware resources are tiled into hardware execution instances and sequentially executed. Panel taken from \citet{spilger2020hxtorch}.
        (C) Example membrane traces during inference of MNIST handwritten digits in the output layer. The ten output activations are sampled at the indicated time for all neurons in parallel.
    }
    \label{fig:ann_operation}
\end{figure}

Extending its application into the realm of non-spiking neural networks, BrainScaleS-2 also allows processing artificial neural networks within its analog core.
This yields several advantages, such as the possibility to process large amounts of input data using convolutional neural networks, and easier multiplexing of the available resources due to the non-time-continuous fashion of the underlying multiply-accumulate operation.
On BrainScaleS-2, input vectors are encoded as arrays of graded spikes, which control the activation time of synapses.
The versatility of the neuron circuits allows them to act as integrators without temporal dynamics, simply accumulating synaptic currents (cf. \cref{ssec:accelerated_emulation,fig:ann_operation}).
The voltage on the membrane capacitances is finally digitized using the \SI{8}{bit} columnar ADC.
The possibility to configure neurons independently allows for a hybrid operating mode with parts of the chip processing ANN layers and other parts processing a spiking network -- a feature unique among accelerators based on analog computation.
A detailed description of this operating mode and the corresponding software interfaces is given by~\citet{weis2020inference} and \citet{spilger2020hxtorch}.

Our software interface enables training of ANNs on BrainScaleS-2 within the PyTorch framework~\citep{paszke2019pytorch} and thereby benefits from well-established gradient-based training methodologies.
We run the forward path of the network on hardware and calculate weight updates on a host computer, assuming an ideal linear model for all computational elements.
The chip is supplied with appropriately-sized MAC operations and executes those in the analog core, the PPU handles simple operations on the results, like pooling or applying activation functions.
Experiment control, like splitting ANNs into simple MAC operations, is usually handled on the host computer.

As an initial proof-of-concept for analog matrix-vector multiplication on BrainScaleS-2, we showcase a simple classifier for the MNIST dataset of handwritten digits~\citep{lecunmnist}.
Using a three-layer convolutional model, we achieve an accuracy of \SI{98.0}{\percent} after training with hardware in-the-loop~\citep{weis2020inference}.
The same model reaches \SI{98.1}{\percent} accuracy on a CPU when discretized to the same \SI{6}{bit} weight resolution.
This operating mode was further used to classify the human activity recognition dataset \citep{spilger2020hxtorch} and to detect atrial fibrillation in electrocardiogram traces in \citet{stradmann2021demonstrating}.

In summary, we have shown that the analog network core of the BrainScaleS-2 system -- in addition to the prevailing spiking operation -- can successfully perform vector-matrix multiplications.
Applying this feature to classical ANNs, competitive classification precision has been reached.
While the current proof-of-concept implementation of this operating mode still carries large potential for future optimizations, interesting hybrid applications combining spiking and non-spiking network layers are already possible with the current hardware generation.

 \section{A Principled Approach to Gradient-Based Parameter Optimisation in Neuromorphic Systems}
\label{ssec:principled_approach_to_learning}
\begin{figure*}[ht!]
\includegraphics{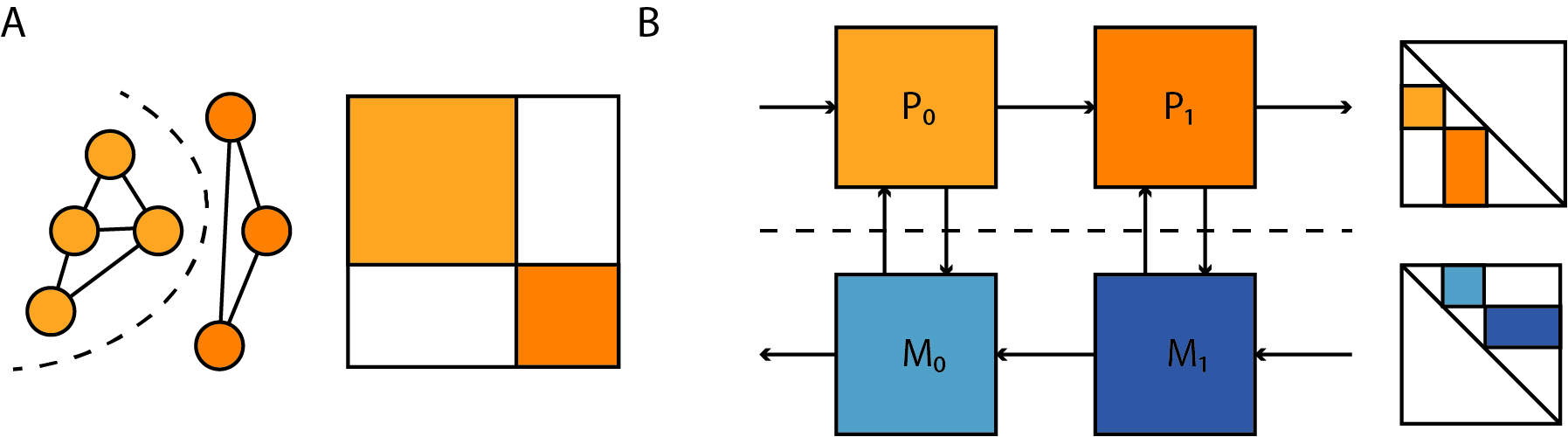}
\caption{(A) A large dynamical system consisting of two decoupled subsystems. It has a block diagonal Jacobian and corresponding decoupled sensitivity equations. (B) Parameter gradient computation in sequentially composed physical systems (orange) can be performed by composing gradient computation in models of the physical systems (blue).}
\label{fig:principled_approach}
\end{figure*}
Given the multitude of approaches to learning and parameter optimization in use in the neuromorphic computing and computational neuroscience community, a natural question arises: Is there a principled way to understand at least gradient-based optimization in parametrized physical systems and (as a particular case) neuromorphic hardware. Here we want to argue that there is such an approach and that it is particularly useful for neuromorphic hardware with complex neuron dynamics and plasticity. As already discussed in the preceding \cref{ssec:gradient_based_learning_approaches}, the key to estimating gradients in a physical system is an appropriate choice of model. The first observation is that most neuromorphic hardware, and in particular the BrainScaleS-2 system, is well described as a hybrid dynamical system \footnote{In the case of fully digital neuromorphic hardware, the situation is more complicated, as they implement a numerical solver for a hybrid dynamical system.}. That is, their dynamics are described by differential equations - the neuron equations, the equations for the correlation traces - together with state transitions - spike-based synaptic input, neuron reset - which happen when jump conditions - a membrane voltage crosses its threshold - are satisfied. The task of estimating gradients in neuromorphic hardware is therefore mainly subsumed under the question of how to compute parameter gradients in a hybrid dynamical system, which is a well-established subject \citep{debacker1964,rozenvasser1967,galan1999parametric,barton2002}. The second observation is that the dynamics in spiking neural networks
decouple, except for spike times, and most of the parameters (the synaptic weights) only enter the state transition functions. Moreover, the jump conditions typically only depend on the state variables of single neuron circuits. These two facts taken together result in simple event-based rules for gradient computation.

More formally a hybrid dynamical system is given by differentiable functions $f^{(s)}(x, p, t)$, labelled by the state $s$ the system is in, which specify the dynamics of the state vector $x$, while in this state:
\begin{align}
\dot{x} = f^{(s)}(x, p, t)
\end{align}
together with jump conditions $j^{(s)}_r(x,p,t) = 0$ and transition equations $x^{+} = T^{(s)}_r(x^-, p, t).$ We use $p$ to indicate a dependence on some number of parameters of both the dynamics, jump condition and transition equations. One simple example would be the Leaky Integrate and Fire neuron model. The state is given by membrane voltage and synaptic input current $x = (V, I)$ of $N$ neurons,
\begin{align}
	\tau_m \dot{V} & = (V_L - V) + R I \\
	\tau_s \dot{I} & = -I
\end{align}
Each jump condition corresponds to the membrane threshold crossing condition of one of the $N$ neurons
\begin{equation}
	V^-_i - (V_T)_i = 0
\end{equation}
and the corresponding transition equation implements the reset of membrane voltage and the jump of the synaptic input current, when neuron $i$ fires:
\begin{align}
	I^+ & = I^- + W e_i \\
	V^+_i & = V_\text{reset},
\end{align}
here $W$ denotes the synaptic weight matrix and $e_i$ is the $i$-th unit vector.The non-zero entries of the Jacobian
\begin{align}
J^{(s)} = \partial_x f^{(s)}(x, p, t)
\end{align}
characterizes which dynamical variables $x_i$ directly couple to each other. A distinguishing feature of all neuromorphic architectures is that even though the overall state space might be large (\numrange{1e5}{1e6} of neuron equations in the case of large scale systems like Loihi, TrueNorth, SpiNNaker and the WaferScale BrainScaleS system), the Jacobian $J^{(s)}$ is sparse and block diagonal. Indeed a system of $N$ Leaky Integrate and Fire neurons has a Jacobian $J$ with diagonal entries
\begin{equation}
\begin{pmatrix}
	-1/\tau_m & R/\tau_m \\
	0         & -1/\tau_s \\ 
\end{pmatrix}
\end{equation}
The Jacobian of a jump condition
\begin{equation}
\partial_{x} j^{(s)}(x,p,t)
\end{equation}
in the case of spiking neural networks is equally sparse, for a Leaky Integrate and Fire neuron model
with threshold $V_T$ in order for the $k$-th neuron to spike
\begin{equation}
j^{k}(x,p,t) = V_k - V_T
\end{equation}
and therefore has only one non-zero entry.

A key observation is that if the state of the system is is known at a time $t_0$ , then the time $t^*$ at which a transition happens is an implicit function of the system's state in a neighborhood close to the transition. One can therefore use the \emph{implicit function} theorem (under certain technical conditions), to compute the parameter derivative $\frac{\mathrm{dt}^*}{\mathrm{dp}}$:
\begin{equation}
\partial_x j \left[ \left( \frac{\partial x}{\partial p} + \partial_t x \right) + \partial_p j + \frac{\mathrm{dt}^*}{\mathrm{dp}} \right] = 0
\end{equation}
In contrast to the time-to-first spike approach \citep{goeltz2019fastNatMI}, this does not require explicit or analytical knowledge of the function $t^\star(x, p)$ and is also applicable to more complex neuron models. In the context of spiking neural networks, this was recognized by \citet{wunderlich2020eventprop} and elaborated in full generality by \citep{pehle2021adjoint}. Concurrent work also introduced this technique to the wider machine learning community \citep{chen2021learning}. This observation is particularly useful in the case of spiking neurons because the jacobian $\partial_x j$ is sparse and therefore results in a sparse coupling of the gradient computation across jumps.

To solve computational tasks, the parameters $p$ of a spiking neural network, such as the weight matrix $W$, need to be optimized according to some (differentiable) loss function. Given such a particular model and task, a constrained optimization problem can be formulated for the parameters $p$, involving an integral over a task-specific loss function $l$
\begin{equation}
\label{eq:loss}
L = \int_0^T l(x,p,t) \mathrm{dt},
\end{equation}
subject to the constraints on the state $x$ given by the equations above. The calculation of the gradient of the loss with respect to the parameters involves the \emph{adjoint equations}
\begin{align}
\lambda'^T = \lambda^T J + \partial_x l,
\end{align}
where the jacobian $J$ of the dynamical system ensures that the computation has the same sparse coupling pattern as the forward equations and $( )'$ indicates the derivative \emph{reverse} in time. By their nature, typical neuromorphic architectures and BrainScaleS-2, in particular, have $O(n)$ parameters, where $n$ is the number of ``neuron'' circuits, which enter the continuous-time evolution. More specifically, in the case of BrainScaleS-2, those are the calibration parameters of the neuron circuits as discussed in \cref{ssec:accelerated_emulation} or rather the model parameters (depending on the viewpoint). A much larger fraction of the parameters $O(n^2)$, namely the synaptic weights, enter only the transition equations.

A similar argument to the one made above for the parameter derivative of the transition times allows one to then relate the adjoint state variables after $\lambda^+$ to the adjoint state variables before the transition $\lambda^-$ and yields an \emph{event-based} rule for gradient accumulation of the parameters that only enter the transition equations (in particular the synaptic weights). This is elaborated more explicitly in \citep{wunderlich2020eventprop, pehle2021adjoint}. The event-based nature of the gradient accumulation and the sparse propagation of error information has immediate consequences for neuromorphic hardware. In particular, it means that only \emph{sparse observations} or measurements are necessary to estimate the gradients successfully, which is a significant advantage over the surrogate gradient approach of \cref{ssec:surrogate_gradient_itl}, which (at least currently) requires dense observations of membrane voltages.

In the context of the \emph{in-the-loop} training paradigm, the general framework sketched here also has attractive consequences. As the numerical implementation and the implemented dynamical system are separate, one can choose appropriate integration methods, such as ones also applicable to multi-compartment neuron models. There is a well-understood way in which numerical implementation of the forward and reverse time dynamics are related (\cite{haier2006geometric} II.3, note that ``adjoint'' there is not used quite in the same sense as here). Since arbitrary loss functions are supported, the hybrid dynamical system used to model the neuromorphic substrate can receive both model and task-specific loss contributions. In particular arbitrary temporally sparse and partial system observations, $\hat{x}_i$ at times $t_i$ can enter a loss term of the form
\begin{equation}
l_m = \sum_i \lvert P x - P \hat{x}_i \rvert^2 \delta(t - t_i),
\end{equation}
where $P$ denotes a linear projection to a state subspace. Such loss terms allow one to account for parameter mismatch and dynamical differences between model and hardware without a need for full access to the system state, which is prohibitive for large neuromorphic systems. Similarly, observations of spikes alone could be used to both fit the model to the neuromorphic substrate and enter the computation of parameter gradients without additional other observations. Moreover, as illustrated in \cref{fig:principled_approach} this extends to situations where one has separate models for physical subsystems -- dynamical models, and adjoint computation of parameter gradients compose as one would expect. This compositionality is useful for the time-multiplexed execution of large feedforward models and the coupling of different parametrized physical substrates, with the goal of task-specific end-to-end optimization. In summary we believe this to be a promising approach to gradient-based optimisation in neuromorphic hardware and intend to apply it to the BrainScaleS-2 architecture presented here.
 
\section{Discussion}
\label{sec:discussion}
We have presented the BrainScaleS-2 system architecture as implemented in a single-chip ASIC with an analog core consisting of \num{512} neurons and 2\textsuperscript{17} synapses, as well as two embedded plasticity and control processors. The system design meets our expectations concerning flexibility and configurability of the analog components and has proven to be a versatile platform for implementing a wide array of tasks across several domains.

The modularity of the architecture, in particular, the neuron circuits, allows for the evaluation of the implemented neuron models on a wide range of parameters as seen in \cref{ssec:faithful_emulation} and even accelerated emulation of multi-compartment neurons \citep{kaiser2021emulating}. The hybrid approach of combining the analog core with flexible digital control and plasticity architecture has enabled many of the experiments reported here. Beyond the immediate practical benefits, it also lays the foundation towards a fully integrated standalone deployment of this neuromorphic architecture, as partially realized by the mobile system \citep{stradmann2021demonstrating}. It is also crucial for the scalable calibration and control of a larger scale architecture, which would use the presented neuromorphic core as a unit of scale.

Another aspect of implementing a large-scale system based on the presented architecture has not been touched upon:
implementing a scalable event-routing architecture and wafer-scale integration. Since the spiking network is emulated completely asynchronously, it has punishing demands on latency and timing jitter.
Future work will have to address this challenge head-on. While scaling to a larger reticle size will be straightforward, accomplishing wafer-scale integration and networking between wafers requires further work and funding.
However, we believe that our experience with the first-generation BrainScaleS wafer-scale system puts us in an excellent position to accomplish a wafer-scale second-generation design \citep{schemmel2010iscas}.
As an intermediate step, we plan to realize a multi-chip system building upon existing BrainScaleS-1 wafer-scale system components.
Interconnectivity will be implemented using \glspl{fpga} based on the EXTOLL network protocol \citep{neuwirth2015scalable,resch2014sustained}.
Concurrently we are working on a low-power inter-chip link, which eventually could also be the basis for on-wafer connectivity. While wafer-scale integration is surely a valuable step to a scale-up
of the system, integrating multiple wafer-scale systems using conventional communication technology will present additional challenges. Furthermore, assembling wafer-scale systems is a time-consuming and difficult task.
We explored techniques such as embedding wafers into printed circuit boards to simplify this task~\citep{zoschkeguettler2017rdlembedding}.

Several digital neuromorphic architectures have been proposed in recent years \citep{furber2012overview, merolla2014million, davies2018loihi, frenkel20180, frenkel2019morphic, pei2019towards, mayr2019spinnaker}. Instead of analog emulation of the neuron and synapse dynamics, these commonly rely on a digital implementation with varying biological faithfulness and flexibility. Other architectures incorporate -- similar to the BrainScaleS platform -- an analog emulation of neuronal and synaptic dynamics \citep{benjamin2014neurogrid, moradi2014eventbased, moradi2018dynaps, neckar2018braindrop, rahimi2020complementary}; a review of earlier approaches were assembled by \citet{indiveri2011neuromorphic}. Some of them are directly inspired by the pioneering work of \citet{mead88silicon} and \citet{mead90neuromorphic} and rely on subthreshold characteristics of transistors. Digital and analog verification methods for the BrainScales-2 system were previously discussed in \citet{grubl2020verification}. A description of the BrainScaleS-2 architecture, event routing, block diagrams, analog-matrix multiplication extensions, on-hardware measurements of matrix multiplication, and the multi-compartment extension, as well as a first application of the
artificial neural network operation mode has been given in \citet{schemmel2020accelerated}.

Considerable progress on learning methods for neuromorphic hardware has been made as well. Here we only highlight
those most closely related to the presented methods. \citet{esser2016convolutional} proposed the use of a pseudo-derivative to replace the derivative of the Heaviside function used for spike threshold detection and also used a forward pass simulating the precision constraints (trinary synapses) of the target hardware and floating-point precision backward pass. Subsequent work applied this training procedure to other neuron models or used other pseudo-derivatives \cite{bellec2018long, neftci2019surrogate}. The in-the-loop training approach was realized on the BrainScaleS platform by \citet{schmitt2017hwitl_nourl} for rate-based models and \citet{cramer2020training} for \glspl{snn} employing individual spikes for information transfer and processing, as summarized in \cref{ssec:surrogate_gradient_itl}.

\citet{petrovici2016stochastic} related stochastically stimulated and recurrently connected populations of LIF neurons in the high-conductance state to Restricted Boltzmann Machines. This relationship suggests finding variational representations of and to sample from high dimensional probability distributions on BrainScaleS-2. The accelerated nature of the system allows one to rapidly produce samples over long periods. Recently such a variational representation was used to represent POVM probability distributions of states in certain quantum systems on BrainScaleS-2 \citep{czischek2019sampling, klassert2021variational, czischek2022spiking}. 

Beyond the experiments presented here, we believe the architecture can serve as a versatile platform for further experiments in multiple directions. One promising direction appears to be to close the loop on two different levels of the system hierarchy, as has been explored as part of the learning-to-learn experiments \citep{bohnstingl2019neuromorphic}. Similarly, the evolving-to-learn framework by \citet{jordan2020evolving} seems predestined for implementation on BrainScaleS-2. The accelerated nature of the system makes it ideal for rapid execution of an inner-loop, involving a bio-inspired learning rule and an additional optimization technique in the outer loop. The flexible architecture for plasticity processing allows one to for example evaluate RNN-based plasticity rules. In another direction online learning rules, such as E-Prop \citep{bellec2020solution} are natural candidates for implementation on BrainScaleS-2. We have found a way to adopt such learning algorithms to our platform, taking into account the constraints on the observability of correlation and neuron membrane traces.

By providing this platform, we hope to establish physical modeling as a versatile tool to the machine learning and
computational neuroscience community. Furthermore, it can not yet be foreseen what deep insights into the collective dynamics of neural networks will be revealed by emulating multiple, interconnected layers of structured neurons, all continuously subjected to complex learning rules and homeostatic processes.

\section*{Conflict of Interest Statement}

The authors declare that the research was conducted in the absence of any commercial or financial relationships that could be construed as a potential conflict of interest.

\section*{Author Contributions}
CP drafted the article and authored all sections except 2.2, 3.1, 3.2.3, 3.3.1, 3.3.2.
CP contributed to the digital design and verification of the BrainScaleS-2 ASIC.
CP conceived and implemented the processor memory interface that enable the experiments reported in 3.3.1 and 3.3.2.
CP implemented the \gls{fpga} I/O interface and Spike router that enabled the experiment reported in section 3.2.2.
CP conceptualized the approach described in \cref{ssec:principled_approach_to_learning}.
SB conceived and implemented parts of the analog neuromorphic core, most notably the analog neuron circuits, evaluated their performance on the \gls{adex} firing patterns and contributed to the software stack.
SB and BC conceptualized, implemented, and evaluated the surrogate-gradient-based learning framework.
BC conceived, implemented, and evaluated the experiments on reservoir computing.
KS conceptualized, implemented, and evaluated all experimental aspects of the accelerated insect navigation and robotics experiments.
KS conceptualized, characterised and implemented the CADC full-custom circuits with the exception of an operational amplifier by HBP collaboration partners from the Sabancı Üniversitesi in Istanbul and another operational amplifier from Matthias Hock \citep{hock14phd}.
KS designed and implemented both the chip-carrier and interface-boards for the BrainScaleS-2 ASIC.
KS contributed additional work to the analog design and tape-out of the BrainScaleS-2 ASIC.
YS contributed to the design and commissioning of the BrainScaleS-2 ASIC and the associated software stack.
JW developed calibration routines, commissioned the non-spiking operating mode, conducted experiments using ANNs and contributed to the software stack.
JK implemented and evaluated the experiments on multi-compartment neurons and contributed to the software stack.
AL ported the insect navigation task to the full-scale system, developed on-chip calibration routines for neuron circuits and contributed to the software stack.
EM provided conceptual and scientific advice.
EM is the lead developer and architect of the BrainScaleS-2 software stack.
JS wrote the initial extended abstract for the article.
JS is the lead designer and architect of the BrainScaleS-2 neuromorphic system.
All authors contributed to and edited the final manuscript.

CP, SB, BC, JK, KS, YS, JW contributed equally.
\section*{Funding}
The research has received funding from
the EC Horizon 2020 Framework Programme under Grant Agreements 720270, 785907 and 945539 (HBP),
the \foreignlanguage{ngerman}{Deutsche Forschungsgemeinschaft} (DFG, German Research Foundation) under Germany’s Excellence Strategy EXC 2181/1-390900948 (the Heidelberg STRUCTURES Excellence Cluster),
the German Federal Ministry of Education and Research under grant number 16ES1127 as part of the \foreignlanguage{ngerman}{\emph{Pilotinnovationswettbewerb "`Energieeffizientes KI-System"'}},
from the Helmholtz Association's Initiative and Networking Fund under project number SO-092 (Advanced Computing Architectures, ACA),
the Manfred Stärk Foundation
and the \foreignlanguage{ngerman}{Lautenschläger-Forschungspreis} 2018 for Karlheinz Meier.

\section*{Acknowledgments}
The authors wish to thank all present and former members of the Electronic Vision(s) research group contributing to the BSS-2 hardware and software system.
Special thanks to Syed Aamir, Simon Friedmann, Andreas Hartel, Vitali Karasenko, Gerd Kiene, Matthias Hock and Andreas Grübl for contributing to the digital and analog hardware design, testing and initial software development, Christian Mauch, Oliver Breitwieser and Philipp Spilger for the production software stack and system operation. We would like to thank Friedemann Zenke and Mihai Petrovici for conceptual advice and inspiration.

We thank S.\ Höppner and S.\ Scholze from the group ``Hochparallele VLSI-Systeme und Neuromikroelektronik'' of C.\ Mayr from TU Dresden for the PLL macro cell  \citep{hoeppner2013compact} and SerDes macros \citep{scholze2012communication} used in the current BSS-2 chip revisions, and Tugba Demirci from EPFL Lausanne for the on-chip fast MADC.

\section*{Data Availability Statement}
The code to access and configure the system is published on \url{github.com/electronicvisions}. Code and data of individual experiments will be made available upon reasonable request.

\bibliographystyle{unsrtnat}
\bibliography{vision}

\end{document}